\documentclass{article}
\DeclareUnicodeCharacter{FF0C}{,}

\usepackage{arxiv}
\usepackage[square,numbers]{natbib}
\usepackage[utf8]{inputenc} 
\usepackage[T1]{fontenc}    
\usepackage{hyperref}       
\usepackage{url}            
\usepackage{booktabs}       
\usepackage{amsfonts}       
\usepackage{nicefrac}       
\usepackage{microtype}      
\usepackage{lipsum}		
\usepackage{graphicx}
\usepackage{doi}
\usepackage{tabularx}
\usepackage{pifont}
\usepackage{graphicx}
\usepackage{subcaption}
\usepackage{amsmath}
\usepackage{mdframed}
\usepackage{longtable}
\newcolumntype{P}[1]{>{\raggedright\arraybackslash}p{#1}}
\usepackage{comment}
\usepackage{csquotes}
\usepackage[table]{xcolor}
\usepackage{multirow}
\usepackage{array}
\usepackage{listings}
\usepackage{svg}
\usepackage{diagbox}
\usepackage[normalem]{ulem} 

\usepackage{soul}
\usepackage{xcolor}

\usepackage{fancyvrb}

\title{Transforming Evidence Synthesis: A Systematic Review of the Evolution of Automated Meta-Analysis in the Age of AI}

\author{ 
	\href{https://orcid.org/0009-0006-6918-4989}{\includegraphics[scale=0.06]{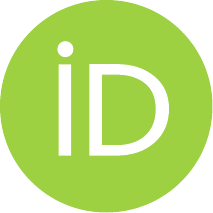}\hspace{1mm}Lingbo Li} \thanks{Corresponding author: L.Li5@massey.ac.nz} \\
	School of Mathematical and Computational Sciences\\
	Massey University\\
	Auckland, New Zealand \\
	\And
	\href{https://orcid.org/0000-0002-9124-2536}{\includegraphics[scale=0.06]{orcid.pdf}\hspace{1mm}Anuradha Mathrani} \\
	School of Mathematical and Computational Sciences\\
	Massey University\\
   Auckland, New Zealand\\
  	\And
\href{https://orcid.org/0000-0001-9416-1435}{\includegraphics[scale=0.06]{orcid.pdf}\hspace{1mm}Teo ~Susnjak} \\
	School of Mathematical and Computational Sciences\\
	Massey University\\
	Auckland, New Zealand \\
}

\renewcommand{\shorttitle}{A Systematic Review of the Evolution of Automated Meta-Analysis}

\begin{document}
\maketitle

\begin{abstract}
Exponential growth in scientific literature has heightened the demand for efficient evidence-based synthesis, driving the rise of the field of Automated Meta-analysis (AMA) powered by natural language processing and machine learning. This PRISMA systematic review introduces a structured framework for assessing the current state of AMA, based on screening 978 papers (2006–2024) and analyzing 54 studies across diverse domains.  Findings reveal a predominant focus on automating data processing (57\%), such as extraction and statistical modeling, while only 17\% address advanced synthesis stages. Just one study (2\%) explored preliminary full-process automation, highlighting a critical gap that limits AMA’s capacity for comprehensive synthesis. Despite recent breakthroughs in large language models (LLMs) and advanced AI, their integration into statistical modeling and higher-order synthesis, such as heterogeneity assessment and bias evaluation, remains underdeveloped. This has constrained AMA’s potential for fully autonomous meta-analysis. From our dataset spanning medical (67\%) and non-medical (33\%) applications, we found that AMA has exhibited distinct implementation patterns and varying degrees of effectiveness in actually improving efficiency, scalability, and reproducibility. 
While automation has enhanced specific meta-analytic tasks, achieving seamless, end-to-end automation remains an open challenge. As AI systems advance in reasoning and contextual understanding, addressing these gaps is now imperative. Future efforts must focus on bridging automation across all meta-analysis stages, refining interpretability, and ensuring methodological robustness to fully realize AMA’s potential for scalable, domain-agnostic synthesis.

\end{abstract}


\section{Introduction}
Automation has become integral to modern life; it is driving efficiencies across industries and is now transforming knowledge intensive domains such as academia \cite{schwab2017fourth}. While businesses have long leveraged automation for operational gains \cite{brynjolfsson2017can}, scholarly research is now accelerating the adoption of AI-driven tools to enhance both efficiency and scalability in evidence synthesis \cite{marshall2019toward}. Systematic Literature Reviews (SLRs) are a cornerstone of academic research and are also among the most resource-intensive academic endeavors, whose workflows stand to be revolutionized by recent advancements in natural language processing (NLP) \cite{kwabena_automated_2023}, machine learning \cite{tsafnat_systematic_2014, NEDELCU2023}, and large language models (LLMs) \cite{khraisha_can_2024} . These technologies are accelerating automation in literature curation, data extraction, and synthesis, and thereby addressing the growing challenge of processing vast and rapidly expanding volume of scientific outputs \cite{van_dinter_automation_2021}.

Meta-analyses (MA) represent a key methodology within the context of SLRs for aggregating quantitative findings \cite{cooper_research_2017, deeks2019meta} for which the current technological advancements present both opportunities and challenges for further automation \cite{van_dinter_automation_2021}. Conducting MAs is resource-intensive, often spanning months or years. With the explosion in the number of papers being published in academic databases, researchers have estimated that the average time  to complete and publish a systematic review requiring five researchers is 67 weeks, with approximate cost of US\$140,000 \cite{borah_analysis_2017}. Moreover, robust MA reviews tend to require an engagement with three to five domain experts to ensure its thoroughness, reliability and accuracy \cite{higgins_cochrane_2019}. Such heavy demands on time, human resource and financial investment pose barriers towards getting timely evidence-based synthesis, particularly in disciplines where rapid and accurate decision-making is essential. Consequently, automation has gained traction across various MA stages to mitigate these constraints. Studies have applied AI-driven techniques to enhance efficiency in literature screening \cite{xiong_machine_2018, chai_research_2021, y_feng_automated_2022}, data extraction \cite{michelson_automating_2014, mutinda_autometa_2022, buchter_systematic_2023, schmidt_exploring_2024}, risk-of-bias assessment \cite{lu_cheng_automated_2021} and heterogeneity reduction \cite{z_rodriguez-hernandez_metagwasmanager_2024}. Despite these gains, automation efforts remain fragmented, with uneven progress across stages, particularly in those requiring complex reasoning and synthesis tasks.

While these advancements contributed towards progress in streamlining various stages of MAs in isolation, no comprehensive undertaking has been made recently to assess the current state of research on the automation of MAs and to situate the existing gaps within the significant and evolving breakthroughs in AI and LLMs, which are increasingly capable of performing complex reasoning \cite{ofori2024towards}. The only dedicated review undertaking a synthesis of AMA, \cite{christopoulou_towards_2023}, focused narrowly on clinical trials, overlooking broader developments in AMA methodologies, while also occurring too early to meaningfully consider the implications of LLMs advancements. Aside from this work, semi-automated meta-analysis (SAMA) has also emerged as an interim solution, shortening MA timelines while maintaining rigor through expert \cite{ajiji_feasibility_2022}. However, SAMA depends on human intervention in key steps, such as study selection and results interpretation, which limits its scalability. Given these shortcomings, a comprehensive review of AMA progress across domains is urgently needed to harness AI’s full potential and address persistent limitations in evidence synthesis automation.

Therefore, this study critically examines the current state of automation in MA research, identifying existing approaches and challenges in preparation for the next wave of AI-driven breakthroughs that are poised further transform the field. It addresses a critical gap by providing the first comprehensive and systematic synthesis of AMA applications across both medical and non-medical domains using a structured analytical framework. Through this analysis, we highlight key challenges and opportunities in AMA and offer insights into its evolving role in quantitative evidence synthesis. Our study therefore makes three meaningful contributions to AMA:

\begin{itemize}
    \item First, it presents a timely systematic and comparative analysis of AMA research progress and applications across medical and non-medical domains to reveal distinct patterns in implementation challenges and opportunities. 
    \item Second, it introduces a structured analytical framework to systematically evaluate the alignment between technological solutions and specific meta-analytical tasks, ensuring more effective automation implementation.
    \item Third, it identifies critical gaps in current AMA capabilities—such as the need for deeper analytical integration and enhanced evidence synthesis—and presents a roadmap for advancement taking the recent AI, and specifically LLM breakthroughs into account.
\end{itemize}

\section{Background}
The following subsection explores the history and evolution of MA, tracing its development from a relatively nascent statistical technique to its current prominence in evidence synthesis across disciplines. The second subsection discusses our analytical framework that informs on technology adoption and task characteristics for conducting the AMA research process. These have helped lay out the research questions for this study.

\subsection{History and Evolution of Meta-Analysis}
MA originated from Glass's pioneering work in the late 1970s, where he developed a statistical framework for synthesizing research findings across educational and psychological studies, formally coining the term "meta-analysis" \cite{glass_meta-analysis_1978}. The methodology expanded significantly into medicine and other scientific domains during the 1980s-90s, particularly for analyzing randomized controlled trials (RCTs). This expansion was driven by the growing demand for evidence-based decision-making, enabling researchers to address contradictory results and overcome limitations of small sample sizes \cite{egger_systematic_2008}. One landmark application in cardiovascular medicine evaluated statin use in reducing cholesterol levels by pooling data from numerous clinical trials to demonstrate clear benefits in lowering heart disease risks, which ultimately provided compelling evidence \cite{collaboration_efficacy_2010}. The field advanced further through more sophisticated statistical models and refined effect size estimation techniques \cite{hedges_statistical_2014}, enhancing the robustness of quantitative synthesis. The development of Preferred Reporting Items for Systematic Reviews and Meta-Analyses (PRISMA) guidelines \cite{takkouche_prisma_2011,page_prisma_2021, page_updating_2021,page_prisma_2021-1}, with its most recent 2020 update, established rigorous reporting standards that minimize bias and improve finding reliability. MA has become instrumental in healthcare research, and as the highest level of evidence synthesis \cite{sackett_evidence-based_1997}, MA provides critical insights for clinical guidelines and public health policies.

However, the exponential growth in published research - exemplified by ScienceDirect (https://www.sciencedirect.com/) with 16 million papers from 2500 journals serving 25 million monthly researchers - has challenged traditional MA approaches. The growing volume of literature has necessitated the development of automation tools to streamline and expedite the review process. Scholars anticipate that automated systematic reviews will revolutionize evidence-based medicine through real-time analysis capabilities and optimized workflows \cite{tsafnat_systematic_2014, van_dinter_automation_2021}. While various software packages (RevMan \footnote{https://training.cochrane.org/online-learning/core-software\#RevMan}, Comprehensive Meta-Analysis \footnote{https://www.meta-analysis.com/}, Stata \footnote{https://www.stata.com/}, SPSS \footnote{https://www.ibm.com/spss}) support MA through features like effect size calculation and heterogeneity assessment, they are better characterized as "computer-assisted" rather than truly automated. For instance, RevMan, despite its user-friendly interface for MA, still requires substantial manual data extraction. 
Similarly, while Comprehensive Meta-Analysis offers advanced statistical modelling, and Stata and SPSS provide flexible analysis capabilities, they all demand significant user intervention and statistical expertise. In addition, the commercial nature and high licensing costs of them limit accessibility for researchers with limited funding.

Recent advancements in AI-driven techniques, such as NLP, machine learning and LLMs, have significantly reduced the time taken during traditional literature synthesis (comprising multiple authors) from many months to few days with use of more efficient automated processes. Despite these advancements, full deployment of AMA remains in development, with current research primarily focused on clinical trials \cite{christopoulou_towards_2023} and SAMA \cite{ajiji_feasibility_2022}. This leaves a significant gap in understanding AMA applications and implications across other fields.

\subsection{SLR Analytical Frameworks}
In a SLR, a robust analytical framework is critical to synthesizing data across studies and drawing meaningful conclusions. The choice of framework depends on various factors, such as the nature of the data, the complexity of the research questions, and the specific domain of interest. Several analytical frameworks support structured analysis and interpretation of findings, including Theory Context Characteristics Methodology (TCCM) \cite{roy_bhattacharjee_brand_2022} to examine theoretical foundations, research contexts, key variables, and methodological approaches; Ability Motivation Opportunity (AMO) \cite{van_dinter_automation_2021} to analyze performance drivers; Unified Theory of Acceptance and Use of Technology (UTAUT) \cite{venkatesh_user_2003} to predict and explain user acceptance and usage behavior of information technology. In this context, Task-Technology Fit (TTF), proposed by Goodhue and Thompson in 1995 \cite{goodhue_task-technology_1995}, offers a valuable analytical lens for assessing how well technology aligns with the tasks it is intended to support, making it particularly suitable for the complexity of AMA. 

TTF posits that the effectiveness of technology adoption and its usage depends significantly on how well the technology supports the specific needs of a given task. It emphasizes on Task Characteristics and Technology Characteristics to make a statement on Task-Technology Fit, that is, to realize if there is a strong fit between technology and task requirements. TTF therefore explains technology adoption (e.g., data locatability, data quality, data accessibility, timeliness, technology reliability, ease of use) by focusing on the alignment between the characteristics of the task and the technology, offering valuable insights for technology design and improvement. The model provides a foundation to understand which technology characteristics would influence the task behaviors and consequently the utilization of that technology for some given purpose. Finally, the utilization would provide a measure of the performance impacts. These impacts could lead to further development of more tools and services, or towards the redesigning of tasks to take better advantage of the technology, or to further embark on training programs to better engage users in using the technologies \cite{goodhue_task-technology_1995}. By applying TTF to AMA, it offers a valuable lens to evaluate the suitability of automation tools across AMA various stages, each with its own set of challenges and needs.
\begin{itemize}
    \item \textbf{Task Complexity with Technology in AMA:} The distinct stages of AMA involve different types of tasks with varying levels of complexity. For instance, in the data extraction phase, automation tools must handle unstructured data from diverse sources, ensuring accuracy in identifying relevant studies and variables. In the synthesis stages, the technology needs to support complex statistical computations while ensuring methodological rigor. In the reporting phase, automation tools must generate clear, interpretable results that comply with reporting standards. At each stage, TTF can decompose tasks and assess how effectively technology meets the specific requirements of each task.
    \item \textbf{Task-Technology Fit Across AMA Stages:} TTF emphasizes the alignment of technology with task characteristics. In AMA, TTF is used to ensure automation technologies remain well-suited to the specific requirements of each task, leading to improved performance and greater user satisfaction.
\end{itemize}

Unlike other analytical frameworks (e.g. TCCM, AMO and UTAUT), 
TTF can delve deeper into the functional match between technology and tasks, which is particularly relevant for complex, multistage process like AMA, where task demands vary significantly at different stages. The application of TTF in AMA allows for a more nuanced understanding of how automation technologies can improve efficiency, accuracy, and user satisfaction. For instance, UTAUT may evaluate whether researchers will intend to use automated tools in MA, TTF assesses whether these tools will improve accuracy and reduce time and labor requirements. Specifically, TTF in AMA helps enable evaluation across three critical dimensions: data quality assessment (reliability of automated extraction), system effectiveness (enhancement of the MA process), and user satisfaction (accessibility across expertise levels). This analytical framework provides structured insights for optimizing AMA tools and advancing task-technology research, ultimately improving both user experience and performance outcomes.

\subsection{Research Questions}
Having laid out the background of history and evolution of MA, this paper applies TTF constructs to better inform on aspects related to AMA deployment, such as the current approaches in use, challenges being faced, future trends and the overall impact on evidence synthesis. We provide a comprehensive review of the development of AMA over the past decade. Our review highlights how various tools are being applied across different disciplines and how they have developed over time to provide a comprehensive understanding of AMA in evidence synthesis. Accordingly, we have posed four research questions that will be addressed in this review. These are:

\begin{itemize}
    \item \textbf{RQ1 (Descriptive)}: What are the current landscape and key characteristics of AMA approaches?
    \item \textbf{RQ2 (Analytical)}: How does our analytical framework illuminate the strengths and limitations of current AMA approaches within each information processing stage?
    \item \textbf{RQ3 (Comparative)}: What are the distinct patterns in AMA implementation, effectiveness, and challenges observed across medical and non-medical domains?
    \item \textbf{RQ4 (Future-Oriented)}: What are the critical gaps and future directions for AMA development, and what obstacles need to be addressed to realize its full potential for evidence synthesis?
\end{itemize}

\section{Methodology}
This section details the methodologies that provide a prelude to the review process and for the presentation of our results. The review follows the PRISMA criteria in providing answers to the four research questions. Next, we introduce PPS to evaluate the alignment between AMA tools and the specific tasks they are designed to support. 

\subsection{PRISMA Process}
Our investigation of automation in MA followed PRISMA guidelines, employing a systematic search strategy. The initial search utilized the string ("meta-analysis" OR "meta analysis") AND ("automation" OR "automated analysis" OR "automated meta-analysis" OR "automatic meta-analysis") across PubMed, Google Scholar, and Scopus databases. This presented a preliminary overview of research activities within the stated field of interest, offering a broad yet comprehensive summary of the general characteristics of MA prevalent in existing literature. Next, we established clear inclusion criteria and practical constraints: (1) published from 2014 to 2024; (2) focus on explicitly MA specific automation tools; (3) full-text availability with detailed technical descriptions and (4) empirical or quantitative evaluations in automation techniques of MA. Table \ref{tab:PRISMA2} details the eligibility criteria. Furthermore, to enhance coverage and overcome potential oversights from database-centric searches, we conducted bidirectional citation chaining “snowball” methods \cite{goodman_snowball_1961,parker_snowball_2019}. This involved both backward tracing (reviewing references cited in the retrieved studies) and forward tracing (identifying later works citing the selected papers) through Scopus and Google Scholar, expanding our temporal scope to 2006-2024. This approach not only expanded coverage by incorporating relevant gray literature and emerging frameworks but also established thematic linkages between foundational methodologies and their contemporary implementations.

\begin{figure}[htbp]
    \centering
    \includegraphics[width=\textwidth]{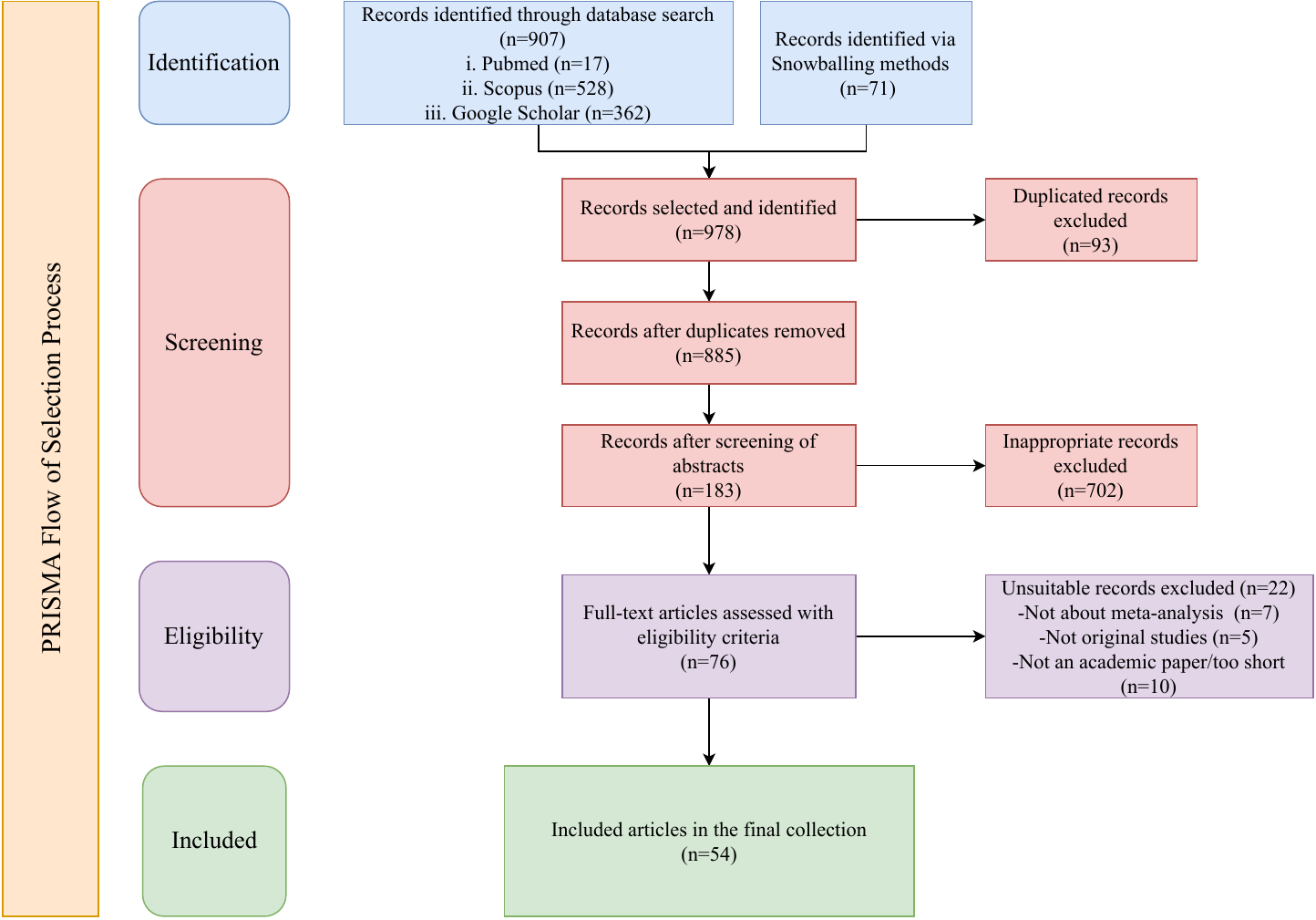}
    \caption{PRISMA workflow}
    \label{fig:PRISMA}
\end{figure}

\begin{figure}[htbp]
    \centering
    \includegraphics[width=0.85\textwidth]{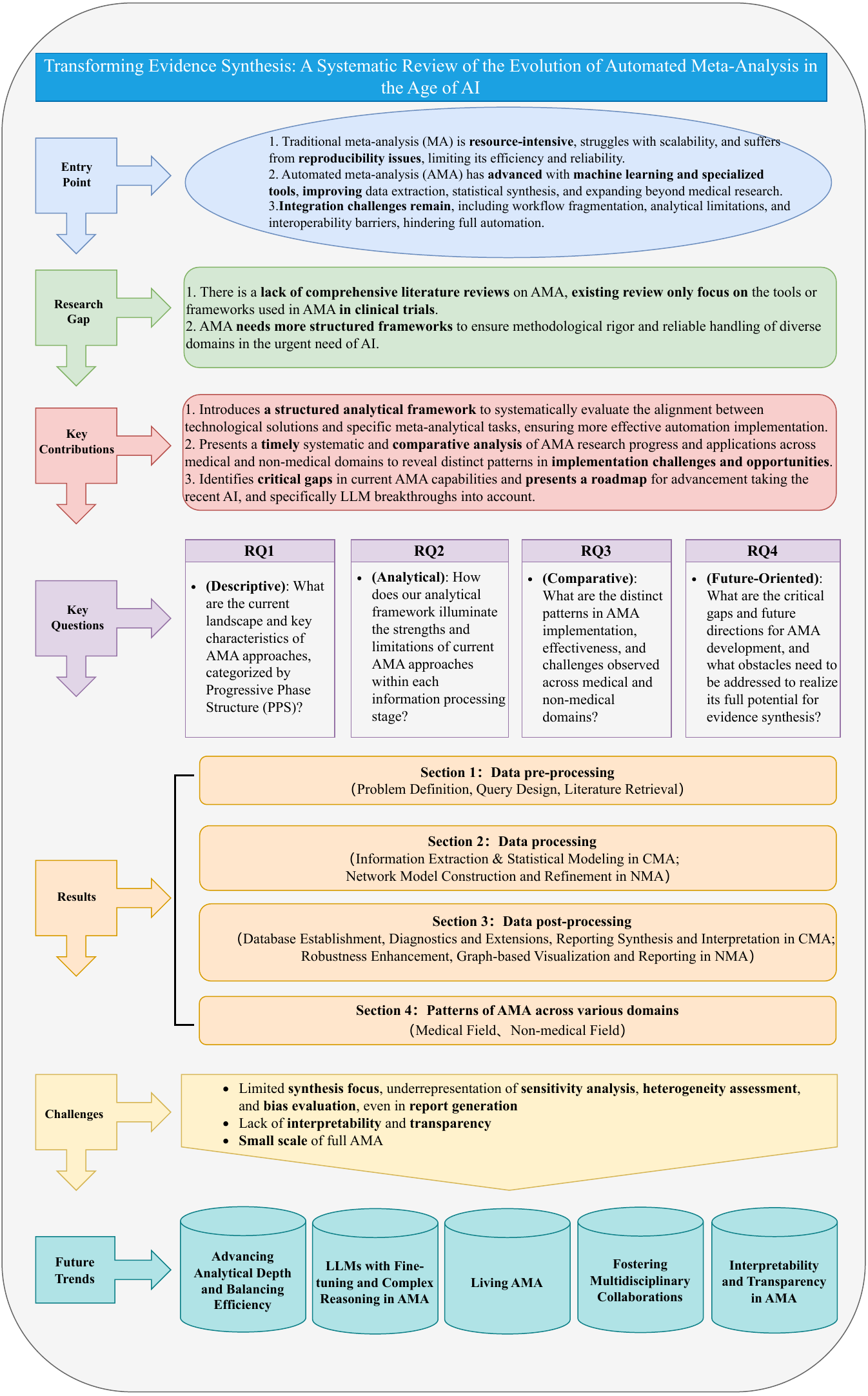}
    \caption{Holistic framework for this review}
    \label{fig:framework}
\end{figure}

The systematic review process, managed through Zotero 7, began with duplicate removal followed by a two-phase screening: title/abstract screening to exclude irrelevant records and full-text review for uncertain cases. Any discrepancies in selection were resolved through consensus discussions among the reviewers. The PRISMA flowchart (shown in Figure \ref{fig:PRISMA}) details the selection process. Data were analyzed and narratively summarized, with descriptive statistics presented in tables or graphs based on each study’s aim. This process identified 978 initial studies (17 PubMed, 528 Scopus, 362 Google Scholar, 71 snowball), which was refined to 54 studies after removing 93 duplicates and excluding 831 studies through screening. The whole visual illustration of our systematic review (shown in Figure \ref{fig:framework}) outlines the key steps, addresses four research questions, highlighting key contributions, current challenges, and future trends in AMA. 

{
\begin{table}[hbt]
    \caption{Criteria for Study Selection Using PRISMA}
    \centering
    \fontsize{8pt}{10pt}\selectfont
    \begin{tabular}{p{2.3cm}p{6cm}p{7cm}}
    \toprule
    \textbf{} & \textbf{Inclusion Criteria} & \textbf{Exclusion Criteria} \\
    \midrule
    \textbf{Study Design} &
    Studies describing or evaluating computational tools or AI-based or machine learning methods applied to at least one stage of the MA process&
    Manual-only MAs without any automated tools or software\\
    \midrule
    \textbf{Technology} &
    Use of automation tools (e.g., text mining, machine learning models, natural language processing) in MA processes &
    Studies without automation applications in literature reviews, synthesis, or data extraction \\
    \midrule
    \textbf{Data Sources} &
    MA studies incorporating datasets from multiple sources (e.g., PubMed, Google Scholar, Scopus)&
    \\
    \midrule
    \textbf{Evaluation Metrics} &
    Studies evaluating the performance of AMA tools, ideally using quantitative metrics (e.g., accuracy, efficiency, time savings). Inclusion is allowed for studies using qualitative metrics if justification for relevance is provided. &
    Studies primarily focused on theoretical discussions of MA without empirical application or evaluation of automation \\
    \midrule
    \textbf{Publication Type} &
    Peer-reviewed journal articles, conference papers, and preprint articles &
    Grey literature, opinion pieces, editorials, or conference abstracts without full methodological detail \\
    \midrule
    \textbf{Language} &
    Articles published in English &
    \\
    \midrule
    \textbf{Time Limit} & 
    All publication dates from 2006 to 2024 were accepted & 
    Duplicates of the same study \\
    \midrule
    \textbf{Outcomes} &
    Studies discussing efficiency, reliability, or scalability improvements in MA due to automation &
    Studies focused on outcomes unrelated to the impact of AMA \\
    \bottomrule
    \end{tabular}
    \label{tab:PRISMA2}
\end{table}
}

\subsection{Progressive Phase Structure in TTF}
AMA streamlines traditional resource-intensive process of MA by integrating automation for data extraction, analysis, and synthesis, enhancing efficiency while reducing human error and statistical expertise requirements There are two primary categories of MA, namely conventional meta-analysis (CMA) for direct comparisons and network meta-analysis (NMA) for integrating both direct and indirect evidence across multiple interventions. Of these, NMA has gained significant traction in clinical medicine, especially when comparing therapeutic options in complex treatment scenarios.

To address RQ1 (What are the current landscape and key characteristics of AMA approaches), we will introduce the Progressive Phase Structure (PPS) to systematically organize automation within MA workflows. Figure 3 illustrates our proposed PPS framework, which categorizes automation processes into three distinct phases across both CMA and NMA:
\begin{itemize}
    \item \textbf{Pre-processing Stage:} Encompasses problem definition, query design, and literature retrieval. NLP, machine learning and LLMs can significantly reduce time spent on these labor-intensive tasks.
    \item \textbf{Processing Stage:} Involves information extraction and statistical modelling (CMA) or network model construction and refinement (NMA). Automated tools leveraging NLP, machine learning, and LLMs help extract required datasets and other relevant information and achieve high efficiency.
    \item \textbf{Post-processing Stage:} Focuses on database establishment, diagnostics, and reporting in CMA, and robustness enhancement and visualization in NMA. Different automation tools can enhance reproducibility through standardized reports and dynamic visualizations, thereby improving transparency.
\end{itemize}

To assess automation effectiveness, we also integrate the TTF model with PPS, providing a structured approach to evaluating alignment between specific MA tasks and available automation tools. This approach systematically deconstructs the automation process into granular components and assesses technological fit at each stage, which operationalizes this alignment by defining:
\begin{itemize}
    \item \textbf{Task characteristics:} Fundamental, well-defined tasks performed at each phase.
    \item \textbf{Technology characteristics:} Capabilities of current automation tools supporting these tasks.
    \item \textbf{Task-Technology fit assessment:} Questions that assess the effectiveness of the technologies, degree of alignment (high/moderate/low), and potential mismatches.
\end{itemize}

\begin{figure}[htbp]
    \centering
    \includegraphics[width=\textwidth]{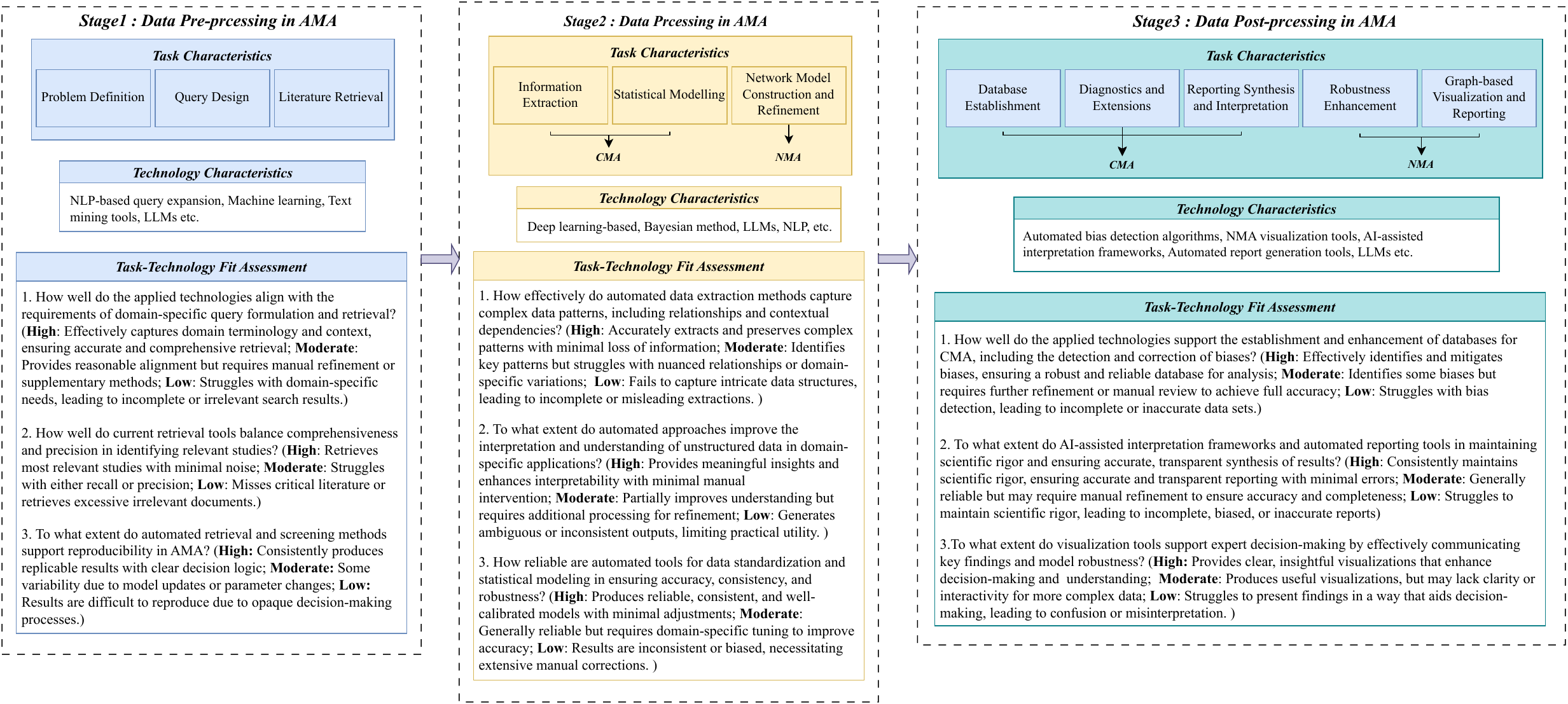}
    \caption{Progressive Phase Structure with TTF model}
    \label{fig:PPS}
\end{figure}

By structuring AMA through PPS and rigorously applying the TTF model (showing in Figure \ref{fig:PPS}), this framework provides a robust methodological foundation for evaluating automation effectiveness in AMA.

\section{Results}
\begin{figure}[htbp]
    \centering
    \includegraphics[width=1\textwidth]{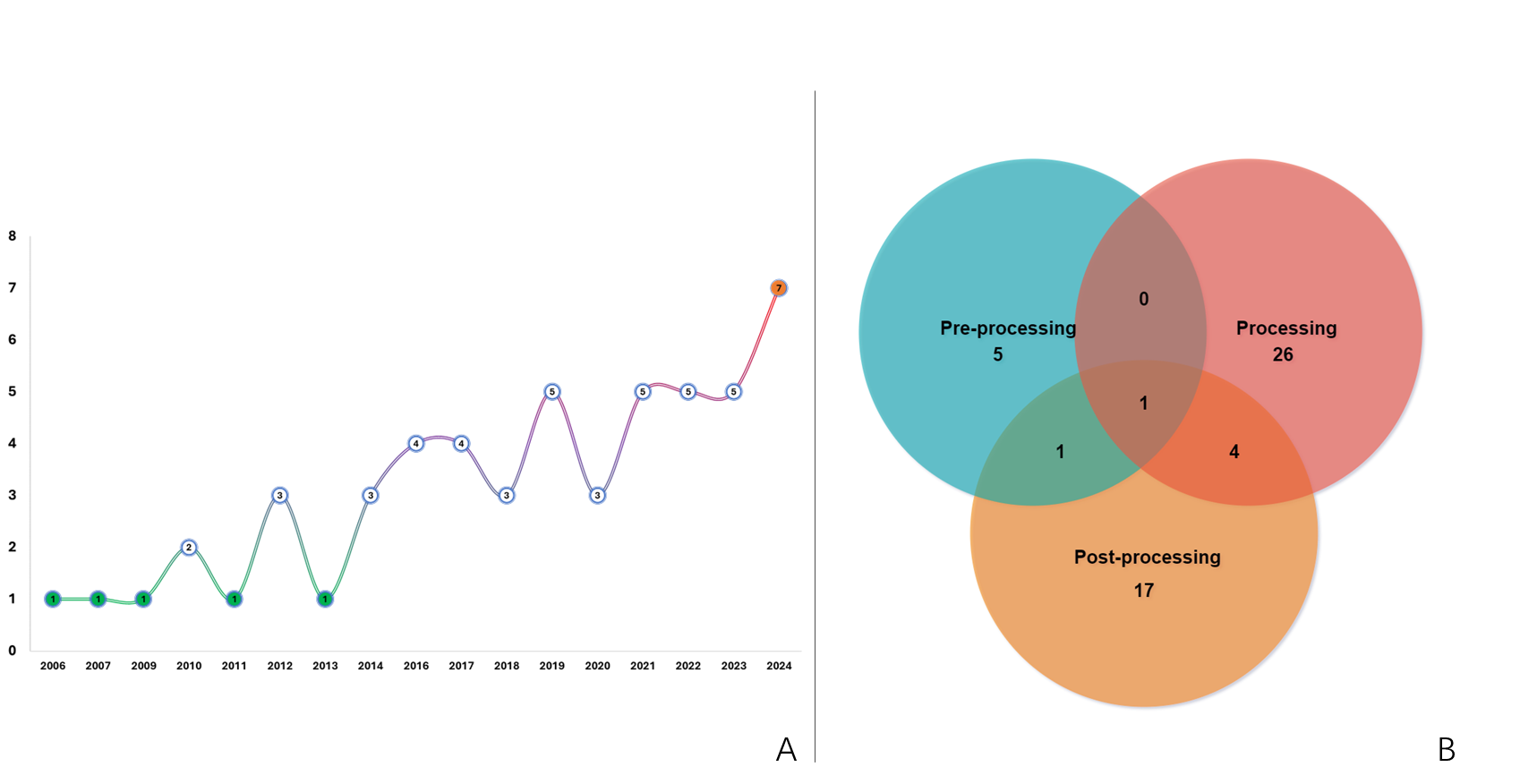}
    \caption{Temporal patterns in AMA publications (A) and proportional discrepancies across different stages (B)}
    \label{fig:timeline-pub}
\end{figure}

Our review identified AMA publications primarily from journals (70\%), conferences (26\%), and preprints (4\%). Figure \ref{fig:timeline-pub}A illustrates the temporal trends showing growth from a single publication in 2006–2009 to seven in 2024. This acceleration, particularly from 2016 onwards, coincides with broader AI advancements and increased availability of computational resources. Despite the growth, the relatively low publication volume indicates AMA remains an emerging field with substantial exploration potential. Besides, analysis of PPS implementation revealed that 89\% of studies focused on automating a specific MA step, while only 11\% addressed multiple stages. Notably, just one study (2\%) attempted full integration across all MA stages, highlighting a significant methodological gap. This indicates that while isolated automation tools have advanced considerably, creating seamless multi-stage workflows remains challenging. Figure \ref{fig:timeline-pub}B shows that processing stage dominates AMA research efforts. This concentration likely stems from technical feasibility and maturity of NLP and machine learning tools for information extraction. As information extraction represents a fundamental prerequisite for all MAs, automation in this area yields substantial efficiency gains. In contrast, the later MA stage involves complex, context-dependent synthesis, which raises further automation challenges, limiting the broader adoption of the system throughout the process.

To address RQ2 (How does our analytical framework illuminate the strengths and limitations of current AMA approaches within each information processing stage?), we provide a comprehensive task breakdown aligned with our analytical framework (refer Figure \ref{fig:PPS}). Automation requirements and success rates vary significantly due to differences in data structure, synthesis models, and computational complexity. Our analysis examines automation strategies and tools employed in both CMA and NMA, identifying distinctive characteristics in each approach. The following subsections detail automation processes across PPS stages within the TTF model.

\subsection{Automation of Data Pre-processing}
Pre-processing in MA comprises problem definition, query design, and literature retrieval, which are all critical for refining datasets for subsequent analysis. The quality of research questions and query design directly influences the relevance and comprehensiveness of retrieved literature. With increasing data volumes, automation has become essential for managing meta-analytic datasets while minimizing bias, as MA performance fundamentally depends on the retrieved literature. Our review findings indicate that existing AMA research in data pre-processing predominantly centers on CMA, with none addressing NMA. Table \ref{tab:pre-processing-ttf} presents a structured evaluation of studies focused on automating this phase, applying TTF to assess alignment between technologies and specific pre-processing tasks. 

Traditional MA requires researchers to craft search strategies that balance breadth and specificity, an inherently complex process that is dependent on researcher expertise \cite{higgins_cochrane_2019}. Automation tools have progressively transformed this landscape. In its initial automation (2017-2018), \cite{cooper_research_2017} introduced MetaBUS, establishing systematic keyword expansion for precise search criteria in human resource management research; \cite{yang_exploration_2018} simplified literature retrieval by organizing PubMed XML files by ID for preliminary screening; \cite{xiong_machine_2018} employed K-means clustering to reduce manual study screening by 87\% in diabetes-atrial fibrillation research, though final selection still required manual review. After that, \cite{anisienia_research_2021} utilized Support Vector Machines (SVMs) to classify research methods within Information Systems literature, addressing limitations of keyword-based approaches. Recent advancements in LLMs have introduced new possibilities for enhancing this process. \cite{wei_chat2brain_2023} developed Chat2Brain, leveraging LLMs to semantically enrich queries for improved study identification; \cite{issaiy_methodological_2024} demonstrated ChatGPT's high sensitivity (95\%) and NPV (99\%) in screening radiology abstracts, reducing workload by 40-83\% despite lower specificity; \cite{luo_evaluating_2024} validated GPT-3.5 Turbo against expert reviews across 24,534 studies, showing high initial accuracy and improved sensitivity in later stages. Despite technological advances, formal automated methods for problem definition and query design remain underexplored. Most research continues to rely on expert-driven approaches for these critical tasks. However, the evolution from simple keyword expansion to sophisticated LLM applications indicates promising directions for comprehensive automation in AMA pre-processing.

{
\fontsize{8pt}{10pt}\selectfont
\begin{longtable}{p{0.5cm} p{3.8cm} p{3.5cm} p{7.0cm}}
\caption{Task-Technology Fit Assessment for Automated Pre-processing in AMA}
\label{tab:pre-processing-ttf}\\ \hline
\toprule
\textbf{Study} & \textbf{Task Characteristics} & \textbf{Technology Characteristics} & \textbf{Task-Technology Fit Assessment} \\
\midrule
\endfirsthead
\toprule
\textbf{Study} & \textbf{Task Characteristics} & \textbf{Technology Characteristics} & \textbf{Task-Technology Fit Assessment} \\
\midrule
\endhead
\multicolumn{4}{r}{\textit{Table continues on the next page...}} \\ 
\midrule
\endfoot
\bottomrule
\endlastfoot

\cite{bosco_metabus_2017} & Query Design, Literature Retrieval & MetaBUS search engine, Boolean operators, Keyword expansion & \textbf{High Fit:} The MetaBUS search engine effectively processes keyword-based queries and expansions. \newline \textbf{Misfit:} Boolean operators may lack flexibility in capturing emerging interdisciplinary nuances. \\
\midrule
\cite{xiong_machine_2018} & Literature Retrieval & Machine learning (K-means, Maximum Entropy Classification) & \textbf{Moderate Fit:} Machine learning classifies and organizes literature based on predefined categories. \newline \textbf{Misfit:} Struggles with complex, non-linear classifications and datasets that do not fit neatly into predefined categories. \\
\midrule
\cite{yang_exploration_2018} & Literature Retrieval & E-utilities programming & \textbf{Low Fit:} Relies on basic PubMed literature retrieval. \newline \textbf{Misfit:} Lacks advanced filtering, limiting its effectiveness for complex searches. \\
\midrule
\cite{anisienia_research_2021} & Literature Retrieval & Deep learning, Support Vector Machines (SVMs) & \textbf{Moderate Fit:} Deep learning and SVMs improve classification accuracy and study grouping. \newline \textbf{Misfit:} Struggles with highly heterogeneous datasets, particularly when research methods or objectives are unclear. \\
\midrule
\cite{wei_chat2brain_2023} & Query Design, Literature Retrieval & NLP-based query expansion, LLMs, Automated database searching & \textbf{High Fit:} LLMs refine domain-specific queries and enhance literature retrieval. \newline \textbf{Misfit:} May underperform in specialized domains requiring expert-level contextual understanding. \\
\midrule
\cite{issaiy_methodological_2024} & Query Design, Literature Retrieval & LLMs (ChatGPT), Automated query expansion & \textbf{High Fit:} LLMs achieved high sensitivity (95\%) and NPV (99\%), reducing workload (40–83\%) compared to general physicians in complex query processing. \newline \textbf{Misfit:} May struggle with highly technical queries requiring deeper contextual adaptation. \\
\midrule
\cite{luo_evaluating_2024} & Query Design, Literature Retrieval & LLMs (GPT-3.5 Turbo) & \textbf{High Fit:} GPT-3.5 Turbo enhances query accuracy and retrieval efficiency against expert reviews. \newline \textbf{Misfit:} May require fine-tuning for specialized or underrepresented research topics. \\
\end{longtable}
}

\subsection{Automation of Data Processing}
Our review highlights the critical role of automation in the data processing phase for both CMA and NMA methodologies. While both approaches aim to enhance meta-analytic efficiency and reliability, they involve distinct automation requirements. CMA prioritizes information extraction and statistical modeling for synthesizing individual study data, whereas NMA focuses on network model construction and refinement, addressing challenges in inconsistency detection and network connectivity assessment. The following subsections provide an in-depth examination of these tasks. 

\subsubsection{Automated Data Processing in CMA}
Following the pre-processing stage, the next critical task in CMA is information extraction and statistical modelling. Information extraction (IE) techniques transform unstructured text into analyzable, structured data—a fundamental prerequisite for CMA. Key subtasks include Named Entity Recognition (NER) for identifying critical variables and Relation Extraction (RE) for determining relationships between entities across research articles. Automated approaches significantly reduce manual data handling requirements. Regarding statistical modelling, automation enhances these processes by improving scalability, reducing manual effort, and ensuring reproducibility. Statistical methodology selection must align with specific data types and research objectives. Some statistical frameworks offer specialized capabilities for particular data structures, while others provide broader applicability across diverse datasets and contexts. Table \ref{tab:processing-ttf} summarizes studies on automating the data processing phase in CMA using TTF model, assessing alignment between technologies and tasks requirements while providing insights into their effectiveness and limitations.

{
\fontsize{8pt}{10pt}\selectfont
\begin{longtable}{p{0.5cm}p{3.8cm} p{3.2cm} p{7.5cm}}
\caption{Task-Technology Fit Assessment for Automated Data Processing in CMA} 
\label{tab:processing-ttf}\\ \hline
\toprule
\textbf{Study} & \textbf{Task Characteristics} & \textbf{Technology Characteristics} & \textbf{Task-Technology Fit Assessment} \\
\midrule
\endfirsthead
\toprule
\textbf{Study} & \textbf{Task Characteristics} & \textbf{Technology Characteristics} & \textbf{Task-Technology Fit Assessment} \\
\midrule
\endhead
\multicolumn{4}{r}{\textit{Table continues on the next page...}} \\ 
\midrule
\endfoot
\bottomrule
\endlastfoot

\cite{michelson_automating_2014} & NER for RCTs & NLP clustering & 
\textbf{Low Fit:} Suitable for early automation but lacks the precision required for complex trial data. \newline \textbf{Misfit:} Low accuracy in initial automation stages makes it unsuitable for large-scale or high-stakes trials without further refinement. \\
\midrule
\cite{boyko_framework_2016} & 
RE in vaccine trials (e.g., dendritic cell vaccination) & 
Machine learning (Random Forest) & 
\textbf{High Fit:} Random Forest excels in hypothesis testing for complex vaccine trials, enabling nuanced predictions. \newline 
\textbf{Misfit:} Requires domain-specific customization, limiting generalizability across medical fields without retraining. \\
\midrule
\cite{neppalli_metaseerstem_2016} & NER in STEM education research & Modular system (MetaSeer.STEM) & \textbf{Moderate Fit:} Modular design allows flexibility across STEM disciplines, improving classifier performance in specific domains. \newline 
\textbf{Misfit:} Lacks fine-tuning for specialized fields (e.g., psychology, economics), reducing accuracy in domain-specific applications. \\
\midrule
\cite{lorenz_automatic_2017} & RE for IPD MA & Machine learning optimization (logic regression combined with simulated annealing) & \textbf{High Fit:} Logic regression models combined with simulated annealing Optimize variable matching in large IPD datasets, reducing human error in MA.  \newline 
\textbf{Misfit:} Scalability issues arise with highly heterogeneous datasets due to computational demands. \\
\midrule
\cite{yang_exploration_2018} & Basic RE for extracting tables from PDFs & SmallPDF tool& 
\textbf{Low Fit:} The SmallPDF tool is effective for simple data extraction but lacks advanced processing capabilities.  \newline 
\textbf{Misfit:} Limited use with poorly formatted or non-standard PDFs. \\
\midrule
\cite{da_devyatkin_towards_2019} & Cancer-related RE in PubMed for tumor types and immunotherapy & 
Syntax-semantic analysis & 
\textbf{Moderate Fit:} Syntax-semantic analysis is well-suited for extracting meaningful relationships in cancer research, aiding in tumor-immunotherapy insights. \newline 
\textbf{Misfit:} Highly specialized for oncology; may not generalize to broader medical applications. \\
\midrule
\cite{pradhan_automatic_2019} & 
Clinical NER from ClinicalTrials.gov & 
EXACT system with web-based data extraction & 
\textbf{High Fit:} The EXACT system achieving 100\% data accuracy and reducing extraction time by 60\% compared to manual methods. Web-based design ensures accessibility for large datasets. \newline 
\textbf{Misfit:} Small validation sample limits generalizability to non-clinical trial datasets. \\
\midrule
\cite{lu_cheng_automated_2021} & 
RE for therapeutic associations & 
NLP processing & 
\textbf{Moderate Fit:} RobotReviewer accelerates MA by extracting therapeutic associations from Cochrane reviews. \newline 
\textbf{Misfit:} Requires manual oversight, limiting full automation, especially for novel therapies. \\
\midrule
\cite{alisa_method_2021} & 
NER+RE in Cytology & 
Rule-based parser, machine learning & 
\textbf{Moderate Fit:} The rule-based parser offers high precision in identifying cytology candidates. \newline 
\textbf{Misfit:} Low recall leads to missed candidates, critical for comprehensive cytology analyses. \\
\midrule
\cite{donoghue_automated_2022} & 
Event‐related potential pattern RE across various clinical conditions & 
Automated text-mining & 
\textbf{Low Fit:} Text-mining is beneficial for aggregating event-related potential data but lacks precision for condition-specific insights. \newline 
\textbf{Misfit:} The text-mining process is highly dependent on data quality; struggles with unstructured or biased sources. \\
\midrule
\cite{mutinda_automatic_2022} & 
Biomedical NER for PICO elements (e.g., breast cancer trials) & 
BERT-based model & 
\textbf{High Fit:} The BERT-based model performs exceptionally well in identifying PICO elements, especially in oncology. The use of pre-trained models allows for fine-tuning with domain-specific datasets, improving oncology trial analysis. \newline 
\textbf{Misfit:} Abstract-based focus limits full-trial context capture; primarily suited for breast cancer studies.  \\
\midrule
\cite{mutinda_autometa_2022} & 
Clinical NER for PICO elements in breast cancer trials & 
BERT-based model & 
\textbf{High Fit:} AUTOMETA enables rapid PICO element extraction in clinical trials, enhancing systematic oncology research. \newline 
\textbf{Misfit:} Relies on abstracts; lacks robustness for heterogeneous clinical trial data. \\
\midrule
\cite{zhang_construction_2022} & 
TCM-related RE & 
VBA-Excel integration with meta-evidence database & 
\textbf{Low Fit:} VBA-Excel integration is effective for field-specific MA in TCM research but is not easily adaptable to other medical or scientific fields. The integration with a meta-evidence database is highly useful for TCM-related insights. \newline 
\textbf{Misfit:} TCM-specific focus restricts cross-disciplinary applicability. \\
\midrule
\cite{kartchner_zero-shot_2023} & 
RE in clinical trial articles & 
LLMs (GPT-3.5 Turbo and GPT-JT) & 
\textbf{High Fit:} Zero-shot models flexibly extract diverse trial relationships; fine-tuning enhances accuracy. \newline 
\textbf{Misfit:} Susceptible to hallucination; requires careful validation. \\
\midrule
\cite{shah-mohammadi_large_2024} & RE in clinical trial articles & NLP for XML content, GPT3.5 & 
\textbf{High Fit:} The RE pipeline effectively supports real-time updates and continuous MA of clinical trial outcomes, identifying relationships between interventions and outcomes. \newline 
\textbf{Misfit:} Struggles with inconsistent data formats across trial reports. \\
\midrule
\cite{yun_automatically_2024} & NER+RE in RCTs& General LLMs & 
\textbf{High Fit:} Effectively extracts binary and continuous outcomes (e.g., mortality) from clinical trials through LLMs. \newline 
\textbf{Misfit:} Struggles with continuous outcomes, leading to inconsistencies in complex measures. \\
\midrule
\cite{wang_metamate_2024} & 
NER+RE for educational systematic reviews & 
Domain-specific LLMs (MetaMate) & 
\textbf{High Fit:} MetaMate leverages domain-specific LLMs for high accuracy in extracting participant and intervention data from educational reviews, streamlining educational MA. \newline 
\textbf{Misfit:} Limited by data scarcity; requires larger datasets for broader applicability. \\
\midrule
\cite{choi_latent_2007} & Integration of heterogeneous biological data with uncertainty & Markov Chain Monte Carlo and Expectation-Maximization algorithms& \textbf{High Fit:} Effectively handles uncertainty and variability in biological data; computationally feasible for broader tasks.\newline 
\textbf{Misfit:} Limited flexibility in incorporating external data beyond biological datasets. \\
\midrule
\cite{marot_moderated_2009} & Statistical modelling with small sample sizes & P-value combination and moderated effect size method & \textbf{High Fit:} Suitable for small sample microarray data with high sensitivity; R package \textit{metaMA} enhances usability. \newline 
\textbf{Misfit:} P-value methods may overestimate significance; effect size methods are more conservative and less sensitive. \\
\midrule
\cite{viechtbauer_conducting_2010} & Statistical Modelling & Random effects and fixed effects models; computationally intensive & \textbf{Moderate Fit:} Flexible for integrating data from multiple sources; supports various data types, but computationally demanding for large-scale datasets. \newline 
\textbf{Misfit:} Lacks real-time analysis of very large datasets; not optimized for speed in high-dimensional data. \\
\midrule
\cite{willer_metal_2010} & Statistical modelling in large-scale genomic MA & C++-based tool & \textbf{High Fit:} High speed and scalability for large genomic datasets, ideal for GWAS MA. \newline 
\textbf{Misfit:} Limited flexibility for non-genomic data types; not adaptable to other biological fields. \\
\midrule
\cite{wang_r_2012} & Statistical modelling in data integration with quality control & Automated pipeline in data normalization and integration & \textbf{High Fit:} Improves data consistency and accuracy across multiple platforms; effective for large-scale gene expression analysis. \newline  
\textbf{Misfit:} Primarily designed for gene expression; less effective for other data types. \\
\midrule
\cite{suurmond_introduction_2017} & Statistical modelling for subgroup and moderator analyses & Excel-based tool & \textbf{Moderate Fit:} Offers an easy-to-use tool for subgroup and moderator analysis in Excel, accessible for a wide range of researchers. \newline \textbf{Misfit:} Lacks advanced features, such as meta-regression with multiple covariates, which would benefit complex analyses with multiple variables. \\
\midrule
\cite{debray_framework_2019} & Statistical modelling of diagnostic and prognostic models & Combination of frequentist and Bayesian methods & \textbf{High Fit:} Comprehensive approach for evaluating predictive models with incomplete data; ideal for diagnostic and prognostic studies. \newline 
\textbf{Misfit:} Complex analysis requiring strong statistical knowledge; may not be accessible for non-expert users. \\
\midrule
\cite{dockes_neuroquery_2020} & Statistical modelling to predict brain activity patterns & NLP-based approach & \textbf{Moderate Fit:} Valuable for neuroimaging analysis; offers predictive insights based on abstract search terms. \newline  
\textbf{Misfit:} Limited to neuroimaging data, not applicable to other fields or generalizable to broader biological analysis. \\
\midrule
\cite{penaloza_towards_2020} & Statistical modelling for combining confidence intervals & Mathematical model for combining proportions and estimating trait prevalence & \textbf{Low Fit:} Offers a method to automate the estimation of prevalence, simplifying some meta-analytic tasks. \newline \textbf{Misfit:} High computational complexity and the need for advanced software make this tool less accessible and suitable for all types of MA tasks. \\
\midrule
\cite{llambrich_amanida_2022} & Statistical Modelling focusing on quality control & P-value and fold change integration & \textbf{Moderate Fit:} Targeted quality control for metabolomics; useful for specialized applications. \newline  \textbf{Misfit:} Limited to metabolomics; not easily adaptable to other data types or broader biological research. \\
\end{longtable}
}

\paragraph{Information Extraction:}Information extraction in CMA has evolved from focused NER techniques to comprehensive relation extraction (RE) systems. \cite{michelson_automating_2014} pioneered structured automation for RCT meta-analysis using NLP to extract key data from abstracts. Building on this foundation, \cite{neppalli_metaseerstem_2016} developed MetaSeer.STEM to automate variable extraction in STEM education research, while \cite{pradhan_automatic_2019} created EXACT, achieving 100\% data accuracy and 60\% reduction in extraction time from ClinicalTrials.gov compared to manual methods. To address limited labeled training data challenges, \cite{alisa_method_2021} combined rule-based and machine learning approaches, using dictionary-based parsing with pre-trained Medical Subject Headings (MeSH) embeddings for entity identification. Deep learning further enhanced extraction capabilities, with \cite{mutinda_automatic_2022} applying BERT-based NER to extract PICO (Participants, Intervention, Control, Outcomes) elements from breast cancer trial abstracts. Their subsequent AUTOMETA system \cite{mutinda_autometa_2022} further streamlined this process, though reliance on abstracts limited its generalizability. Moving beyond entity identification, RE techniques established associations between key variables. \cite{boyko_framework_2016} combined random forest classification with RE for constructing causal hypotheses in dendritic cell vaccination research. For structured data extraction, \cite{yang_exploration_2018} developed methodology for automating tabular data extraction from PDFs, while \cite{da_devyatkin_towards_2019} applied RE to identify associations between tumor types and immunotherapy methods. Field-specific adaptations emerged with \cite{lu_cheng_automated_2021} utilizing RobotReviewer \cite{marshall_robotreviewer_2016} to extract therapeutic associations, and \cite{lorenz_automatic_2017} addressing IPD meta-analysis challenges through logic regression models for variable matching. \cite{zhang_construction_2022} developed specialized RE for Traditional Chinese Medicine splenogastric disease research, while \cite{donoghue_automated_2022} employed text-mining for meta-analyzing event-related potential studies across cognitive domains.

Recent innovations have leveraged LLMs for comprehensive information extraction. \cite{wang_metamate_2024} introduced MetaMate, the first LLM-powered extraction tool in education, identifying 20 PICO-related elements. \cite{yun_automatically_2024} focused on LLM annotation of ICO triplets in RCT reports, though continuous outcomes and overlapping measures remained challenging. For relation extraction, \cite{kartchner_zero-shot_2023} evaluated GPT-3.5 Turbo and GPT-JT for extracting study characteristics and risk of bias, identifying hallucination tendencies despite strong zero-shot performance. \cite{shah-mohammadi_large_2024} developed a fully automated pipeline linking interventions and outcomes through XML content extraction, enabling real-time MA updates despite format inconsistencies in clinical trial data. Despite significant advancements, challenges persist in handling continuous numerical outcomes, ensuring consistency across diverse research domains, and mitigating hallucination risks in LLM-based systems. Addressing these limitations remains critical for further advancing automated information extraction methodologies in CMA.

\paragraph{Statistical Modelling:}Statistical modelling advancements have played a pivotal role in enhancing CMA efficiency for heterogeneous datasets. Early approaches focused on probabilistic modelling to manage data variability and uncertainty. \cite{choi_latent_2007} introduced a mixture probabilistic model employing Markov Chain Monte Carlo and Expectation-Maximization algorithms for automated biological data analysis—computationally efficient but limited in external data integration. \cite{marot_moderated_2009} extended this framework with a moderated effect size combination method for microarray MA, particularly beneficial for small sample sizes, finding that P-value combination techniques enhanced sensitivity and gene ranking accuracy. As CMA expanded to high-dimensional data, researchers developed more scalable frameworks. \cite{viechtbauer_conducting_2010} created the \textit{metafor} package, providing flexible modelling for both fixed and random effects models with minimal user intervention. For genomic applications, \cite{willer_metal_2010} introduced METAL, optimized for rapid combination of summary statistics across multiple genetic studies, offering superior speed and scalability though limited to genomic contexts.

Addressing the need for accessible tools, \cite{suurmond_introduction_2017} developed Meta-Essentials, an Excel-based solution focused on user-friendliness for foundational meta-analytic tasks, though lacking advanced functionalities like multi-covariate meta-regression. Quality control advancements emerged with \cite{wang_r_2012} introducing \textit{MetaOmics}, an R-based framework integrating gene expression data across platforms with normalization measures to enhance consistency. \cite{llambrich_amanida_2022} later extended these principles to metabolomics with AMANIDA, a visualization-based tool for integrating P-values and fold changes in metabolic studies. Recent developments have focused on quantitative prediction and hypothesis formulation. \cite{debray_framework_2019} created \textit{metamisc}, integrating frequentist and Bayesian methods to evaluate prediction model performance for binary and time-to-event outcomes while addressing incomplete data challenges. In neuroimaging, \cite{dockes_neuroquery_2020} developed NeuroQuery, aggregating neuroimaging data to predict brain activity patterns from abstract search terms, demonstrating automated hypothesis generation potential. Novel statistical approaches have emerged to address specific meta-analytic challenges. \cite{penaloza_towards_2020} proposed methods for combining confidence intervals to compute proportions across populations, offering more accurate prevalence estimations despite computational complexity. These developments illustrate the progressive refinement from early probabilistic models to advanced statistical frameworks and predictive analytics. Future work will likely focus on enhancing automation, improving integration across diverse data sources, and developing more adaptive meta-analytical models responsive to heterogeneous research contexts.

\subsubsection{Automated Data Processing in NMA}
In NMA, constructing and refining network models represents a critical challenge distinct from CMA. While CMA primarily extracts information and develops statistical models, NMA must integrate both direct and indirect evidence across multiple interventions through complex network structures. Automation in this domain enhances model consistency, computational efficiency, and reduces manual intervention. Table \ref{tab:processing-NMA} provides an overview of the included studies in NMA through TTF model.

Dr. Van Valkenhoef pioneered NMA automation by developing a Bayesian consistency model generation framework \cite{van_valkenhoef_automating_2012} that transformed what was previously a manual process requiring subjective parameter decisions. Building on this foundation, they introduced automated node-splitting to quantify discrepancies between direct and indirect evidence \cite{van_valkenhoef_automated_2016}. Further advancing the field, \cite{thom_automated_2019} implemented a graph-theory-based method to assess network connectivity and evaluate indirect comparison reliability. These sequential innovations have significantly reduced subjectivity while enabling analysis of increasingly complex treatment networks.

{
\begin{table}[!ht]
\caption{Task-Technology Fit Assessment for Automated Data Processing in NMA}
\centering
\fontsize{8pt}{10pt}\selectfont
\begin{tabular}{p{0.5cm} p{3cm} p{3.5cm} p{8cm}}
\toprule
\textbf{Study} & \textbf{Task Characteristics} & \textbf{Technology Characteristics} & \textbf{Task-Technology Fit Assessment} \\
\midrule
\cite{van_valkenhoef_automating_2012} & Construction and refinement of network models & Bayesian models for complex network structures & \textbf{Moderate Fit:} Enables analysis of large treatment networks; improves efficiency by reducing subjectivity in model setup. \newline \textbf{Misfit:} High computational demand, particularly for large and dense networks. \\
\midrule
\cite{van_valkenhoef_automated_2016} & Construction and refinement of network models (automated detection of inconsistencies) & Node-splitting method for identifying discrepancies between direct and indirect evidence & \textbf{High Fit:} Enhances reliability by detecting inconsistencies; reduces need for manual verification. \newline \textbf{Misfit:} Identifies inconsistencies but does not provide mechanisms to resolve them. \\
\midrule
\cite{thom_automated_2019} & Construction and refinement of network models (network connectivity analysis) & Graph-theory-based & \textbf{Moderate Fit:} Reduces manual workload in connectivity assessment; improves confidence in indirect comparisons. \newline \textbf{Misfit:} Primarily focuses on network structure; lacks adaptability for other NMA tasks. \\
\bottomrule
\end{tabular}
\label{tab:processing-NMA}
\end{table}
}

\subsection{Automation of Data Post-Processing}
Having examined automation in data pre-processing and processing for both CMA and NMA, we now focus on data post-processing, a critical phase involving result refinement and synthesis to ensure reporting accuracy and clarity. This increasingly important area of AMA research enhances analytical precision and advances evidence synthesis. The following subsections will provide a detailed characteristics of CMA and NMA post-processing automation. 

\subsubsection{Automated Data Post-Processing in CMA }
Our examination categorizes CMA automated post-processing into three domains: (1) database establishment for structured data organization; (2) diagnostics and extensions for bias or heterogeneity assessment; and (3) reporting synthesis and result interpretation for standardized findings presentation. Table \ref{tab:postprocessing-ttf} presents the TTF alignment for CMA post-processing studies.

{
\fontsize{8pt}{10pt}\selectfont
\begin{longtable}{p{0.5cm}p{3cm}p{3.2cm}p{7.5cm}} 
\caption{Task-Technology Fit Assessment for Automated Data Post-processing in CMA} 
\label{tab:postprocessing-ttf}\\ \hline
\toprule
\textbf{Study} & \textbf{Task Characteristics} & \textbf{Technology Characteristics} & \textbf{Task-Technology Fit Assessment} \\
\midrule
\endfirsthead
\toprule
\textbf{Study} & \textbf{Task Characteristics} & \textbf{Technology Characteristics} & \textbf{Task-Technology Fit Assessment} \\
\midrule
\endhead
\multicolumn{4}{r}{\textit{Table continues on the next page...}} \\ 
\midrule
\endfoot
\bottomrule
\endlastfoot

\cite{yarkoni_large-scale_2011} & Database Establishment & Text mining, machine learning & \textbf{Low Fit:} Automated extraction of brain-cognitive links to handle large-scale brain-mapping database. \newline \textbf{Misfit:} Limited by lexical coding approaches and inconsistencies in brain coordinate reporting, reducing mapping accuracy and reliability. \\
\midrule
\cite{feichtinger_cancerma_2012} & Database Establishment & R and MySQL& \textbf{High Fit:} Efficiently handles large datasets and automates the analysis of gene expression, providing robust insights for cancer biomarker identification. \newline \textbf{Misfit:} Excludes newer cancer types and lacks cutting-edge methodologies, restricting its applicability to recent research. \\
\midrule
\cite{j_feichtinger_cancerest_2014} & Database Establishment & MySQL and web technologies & \textbf{High Fit:} Web-based platform enhances accessibility and simplifies interaction with large genomic datasets. \newline \textbf{Misfit:} Limited to specific tissue types and lacks updates for new biomarker data, reducing applicability across all tissues. \\
\midrule
\cite{michelson_automating_2014} & Result Synthesis and Interpretation & Greedy clustering& \textbf{Low Fit:} Automates trial data aggregation, reducing manual effort. \newline \textbf{Misfit:} Relies on basic clustering, missing advanced techniques for nuanced analysis such as meta-regression or subgroup analyses for nuanced interpretation. \\
\midrule
\cite{shashirekha_shinymde_2016} & Database Establishment & R & \textbf{High Fit:} Enables non-experts to conduct complex gene expression analyses across platforms, enhancing accessibility.  \newline \textbf{Misfit:} Limited preprocessing methods and dataset availability may exclude nuanced or large-scale genomic studies. \\
\midrule
\cite{craig_bridging_2017} & Diagnostics and extensions in extracting effect sizes & Web crawlers and NLP & \textbf{Moderate Fit:} Semantic search capabilities are enhanced through NLP, improving the extraction of relevant research from large datasets. \newline \textbf{Misfit:} System complexity limits accessibility, especially for users without expertise in web crawlers and NLP. \\
\midrule
\cite{yang_exploration_2018} & Result Synthesis and Interpretation & R-\textit{Meta} package & \textbf{Moderate Fit:} Streamlined the MA process by providing a user-friendly framework for applying various statistical methods. Facilitates integration of multiple datasets efficiently. \newline \textbf{Misfit:} Lacks advanced features such as multi-covariate meta-regression, limiting complex analyses. \\
\midrule
\cite{hu_metacyto_2018} & Result Synthesis and Interpretation & Clustering methods without parameter tuning & \textbf{Moderate Fit:} Efficiently identifies common cell populations efficiently for cytometry studies. \newline \textbf{Misfit:} Limited in detecting rare cell populations, potentially missing critical insights.  \\
\midrule
\cite{lu_cheng_automated_2021} & Diagnostics and extensions in detecting bias & Causal inference techniques & \textbf{Moderate Fit:} Improves RCT validity by controlling for biases and confounding. \newline \textbf{Misfit:} Assumes single-ignorability and is restricted to RCTs, reducing versatility for observational studies. \\
\midrule
\cite{de_bruijne_text2brain_2021} & Result Synthesis and Interpretation & Deep learning (neural network) & \textbf{High Fit:} Provides a powerful tool for large-scale neuroimaging research, synthesizing brain activation maps with improved accuracy over traditional methods. \newline \textbf{Misfit:} Struggles with ambiguous text queries, potentially leading to mapping inaccuracies. \\
\midrule
\cite{sabates_cogtale_2021} & Reporting synthesis and result interpretation & Web-based platform & \textbf{Moderate Fit:} Provides a user-friendly semi-automated platform that supports cognitive treatment analysis and evidence-based recommendations. \newline \textbf{Misfit:} Requires manual data extraction, limiting full automation and increasing human error risk. \\
\midrule
\cite{wei_chat2brain_2023} & Reporting synthesis and result interpretation & LLMs; integrates with Text2Brain & \textbf{Moderate Fit:} LLMs enhance query flexibility and improve alignment with brain activation distributions, addressing Text2Brain’s semantic redundancy. \newline \textbf{Misfit:} Still constrained by brain coordinate complexity and ambiguous queries, affecting accuracy. \\
\midrule
\cite{finnigan_retrobiocat_2023} & Database Establishment & Python Flask web server & \textbf{Moderate Fit:} Facilitates real-time data updating and provides an accessible biocatalysis analysis platform. \newline \textbf{Misfit:} Limited to biocatalysis, reducing applicability to broader biological and chemical domains. \\
\midrule
\cite{z_rodriguez-hernandez_metagwasmanager_2024} & Diagnostics and extensions in reducing heterogeneity& R, Bash, and Python & \textbf{High Fit:} Standardizes workflow and reduces heterogeneity in large GWAS meta-analyses. \newline \textbf{Misfit:} Lacks details on functionalities and customization options, potentially affecting adoption in specialized GWAS studies.  \\
\end{longtable}
}

\paragraph{Database Establishment:}In CMA, specialized databases are crucial for research automation. \cite{yarkoni_large-scale_2011} pioneered an automated brain-mapping framework integrating text mining and machine learning, linking neural activity with cognitive states. While expediting the mapping process, its lexical coding approach and inconsistent brain coordinate reporting may generate false positives. Feichtinger et al. developed two significant tools: CancerMA \cite{feichtinger_cancerma_2012}, a pipeline for analyzing gene expression across 80 cancer microarray datasets to identify biomarkers, and CancerEST \cite{j_feichtinger_cancerest_2014}, a web-based tool for identifying cancer markers using expressed sequence tags from 36 tissue types, advancing personalized treatment research. Moreover, \cite{shashirekha_shinymde_2016} created ShinyMDE, a flexible tool integrating gene expression data from multiple platforms to detect differential expression, providing a user-friendly solution for cross-platform meta-analysis. More recently, \cite{finnigan_retrobiocat_2023} developed RetroBioCat, an integrated database for biocatalysis data that addresses paper-based format limitations through continuous updating and improved accessibility.

\paragraph{Diagnostics and Extensions:}Diagnostics and extensions enhance CMA accuracy and robustness through sensitivity analysis, heterogeneity assessment, and bias evaluation, improving meta-analytic finding generalizability. \cite{craig_bridging_2017} pioneered automated effect size extraction using web crawlers and NLP techniques integrated with NPDS infrastructure, incorporating semantic query expansion to curate relevant research data while reducing manual effort. \cite{lu_cheng_automated_2021} advanced the field by integrating causal inference techniques into CMA, addressing hidden confounders in clinical data, which is the first systematic approach to account for bias in automated meta-analytic workflows. This framework improved result quality and accuracy across medical and clinical studies. Building on these foundations, \cite{z_rodriguez-hernandez_metagwasmanager_2024} developed metaGWASmanager, a toolbox optimizing genome-wide association studies in large-scale MA consortia. By standardizing methodologies across central analysis groups and study analysts, this tool minimized heterogeneity from differing data preparation techniques, enhancing reliability and reproducibility of genetic findings.

\paragraph{Result Synthesis and Interpretation:}
Result synthesis and interpretation are essential for transparency and reproducibility in CMA, enhancing interpretability and supporting evidence-based decisions. \cite{michelson_automating_2014} pioneered this area using greedy clustering with Paule-Mandel random effects models, establishing a foundation for automated synthesis. \cite{yang_exploration_2018} advanced this work by applying the R-\textit{Meta} package for more precise result synthesis. \cite{hu_metacyto_2018} developed MetaCyto, automating meta-analysis of cytometry data without parameter tuning requirements, while \cite{sabates_cogtale_2021} created CogTale, a semi-automated platform for cognitive-oriented treatments in older adults that generates treatment effect estimates despite requiring manual data extraction. \cite{de_bruijne_text2brain_2021} introduced Text2Brain, a neural network tool for coordinate-based meta-analysis of neuroimaging studies, synthesizing brain activation maps despite limitations in complex query interpretation. \cite{wei_chat2brain_2023} subsequently enhanced this approach with Chat2Brain, applying LLMs to process semantic queries, improving activation pattern accuracy and addressing language ambiguities.

\subsubsection{Automated Data Post-processing in NMA }
In contrast to CMA, automated post-processing in NMA focuses on two primary areas: (1) robustness enhancement to ensure network model validity and stability; and (2) graph-based visualization and reporting to facilitate complex treatment network interpretation. Table \ref{tab:postprocessing-NMA} presents studies relevant to NMA post-processing.

{
\begin{table}[!ht]
\caption{Task-Technology Fit Assessment for Automated Data Post-processing in NMA}
\centering
\fontsize{8pt}{10pt}\selectfont
\begin{tabular}{p{0.5cm} p{3cm} p{3.2cm} p{8.2cm}}
\toprule
\textbf{Study} & \textbf{Task Characteristics} & \textbf{Technology Characteristics} & \textbf{Task-Technology Fit Assessment} \\
\midrule

\cite{van_valkenhoef_addis_2013} & Graph-based visualization and reporting & ADDIS tool & \textbf{Moderate Fit:} Makes NMA more accessible to users without deep statistical expertise. \newline \textbf{Misfit:} Limited advanced analytical features for experts; mainly designed for simpler analyses. \\
\midrule
\cite{neupane_network_2014} & Robustness enhancement & Comparison of R packages: \textit{gemtc}, \textit{pcnetmeta}, and \textit{netmeta}; focuses on Bayesian vs. frequentist methods. & \textbf{High Fit:} Helps researchers select tools based on their specific needs (e.g., statistical method preference, flexibility). \newline \textbf{Misfit:} Limited guidance for researchers without statistical expertise, as some tools are highly technical. \\
\midrule
\cite{owen_metainsight_2019} & Graph-based visualization and reporting & MetaInsight graphical interface& \textbf{Moderate Fit:} Provides an easy-to-use interface that facilitates NMA for those without statistical expertise. \newline \textbf{Misfit:} Lacks flexibility for users needing advanced statistical modeling beyond basic workflows. \\
\midrule
\cite{nikolakopoulou_cinema_2020} & Robustness enhancement focusing on study bias, uncertainty, heterogeneity, etc. & CINeMA framework & \textbf{High Fit:} Provides an in-depth, structured approach to assessing NMA credibility, enhancing reliability and robustness. \newline \textbf{Misfit:} Can be resource-intensive, requiring substantial data input and expert judgment for full evaluation. \\
\midrule
\cite{chiocchia_semi-automated_2023} & Robustness enhancement focuses on addressing missing data bias & ROB-MEN tool& \textbf{Moderate Fit:} Enhances NMA robustness by addressing missing data, a common issue in systematic reviews. \newline \textbf{Misfit:} Focuses narrowly on missing data, excluding other biases, and may require expert interpretation. \\
\midrule
\cite{y_liu_gentle_2023} & Graph-based visualization and reporting & \textit{BUGSnet} R package & \textbf{High Fit:} Makes Bayesian NMA more accessible in social science, allowing for complex data handling and incorporating prior knowledge. \newline \textbf{Misfit:} Requires R programming knowledge, limiting access for those without coding expertise. \\
\midrule
\cite{reason_artificial_2024} & Robustness enhancement & LLMs (GPT-4) & \textbf{High Fit:} Reduces time and effort involved in data extraction and script writing, improving efficiency and scalability in NMA. \newline \textbf{Misfit:} Potential for errors in data extraction and interpretation due to limitations of LLMs, impacting overall accuracy. \\
\bottomrule
\end{tabular}
\label{tab:postprocessing-NMA}
\end{table}
}

\paragraph{Robustness Enhancement:}
Specialized tools and frameworks have significantly enhanced NMA robustness. \cite{neupane_network_2014} compared three key R packages (\textit{gemtc}, \textit{pcnetmeta}, and \textit{netmeta}), evaluating their usability, flexibility, and computational efficiency to guide researchers in package selection. \cite{nikolakopoulou_cinema_2020} advanced credibility assessment through CINeMA, a framework systematically evaluating factors affecting NMA reliability, including study bias, indirectness, uncertainty, heterogeneity, and inconsistency. Complementing this work, \cite{chiocchia_semi-automated_2023} developed ROB-MEN to address bias from missing data, further strengthening analysis reliability. Recently, \cite{reason_artificial_2024} demonstrated that LLMs (GPT-4) can efficiently extract data for NMA, reducing manual effort and accelerating analysis. This integration of AI technologies promises enhanced efficiency and accuracy for large-scale NMA.

\paragraph{Graph-based Visualization and Reporting:}
Graph-based visualization is essential for interpreting complex treatment networks in NMA. \cite{van_valkenhoef_addis_2013} pioneered this area with ADDIS, a specialized system facilitating clinical trial data input, treatment comparison, and conclusion derivation, making NMA more accessible for healthcare policy and decision-making. \cite{owen_metainsight_2019} later developed MetaInsight, an interactive tool with a graphical interface that guides users through the NMA process without requiring deep statistical expertise. This tool democratized access to complex statistical methods, enabling researchers across disciplines to perform NMA more efficiently. \cite{y_liu_gentle_2023} expanded Bayesian NMA into psychology and social sciences through their automated R package \textit{BUGSnet}. This approach provides more accurate treatment effect estimates by incorporating prior information and handling complex data structures, increasing Bayesian NMA accessibility in fields where such methods were previously uncommon.

\subsection{Patterns of AMA Across Domains}
The PPS with TTF model provides a comprehensive framework for AMA by automating each process step. To answer RQ3 (What are the distinct patterns in AMA implementation, effectiveness, and challenges observed across medical and non-medical domains?), our analysis revealed significant domain-specific variations.

A primary distinction lies in data characteristics. Medical domains utilize standardized, structured data from clinical trials, healthcare records, and standardized literature. This creates a strong task-technology fit for automated tools, such as NLP, machine learning, LLMs etc. which efficiently process consistent terminology with minimal human intervention. Conversely, non-medical fields (social sciences, management, education, STEM) present heterogeneous, less structured data with varied reporting styles and terminologies, creating a misfit for current text mining and automated tools primarily designed for structured data. Methodological traditions further differentiate these domains. In medical domain, established protocols, statistical methods, and data collection guidelines support automation through predictable, uniform data formats. Non-medical fields lack such standardization, with diverse research methods and reporting approaches complicating automated tool application and exacerbating the task-technology misfit.

\begin{figure}[htbp]
    \centering
    \includegraphics[width=1.0\textwidth]{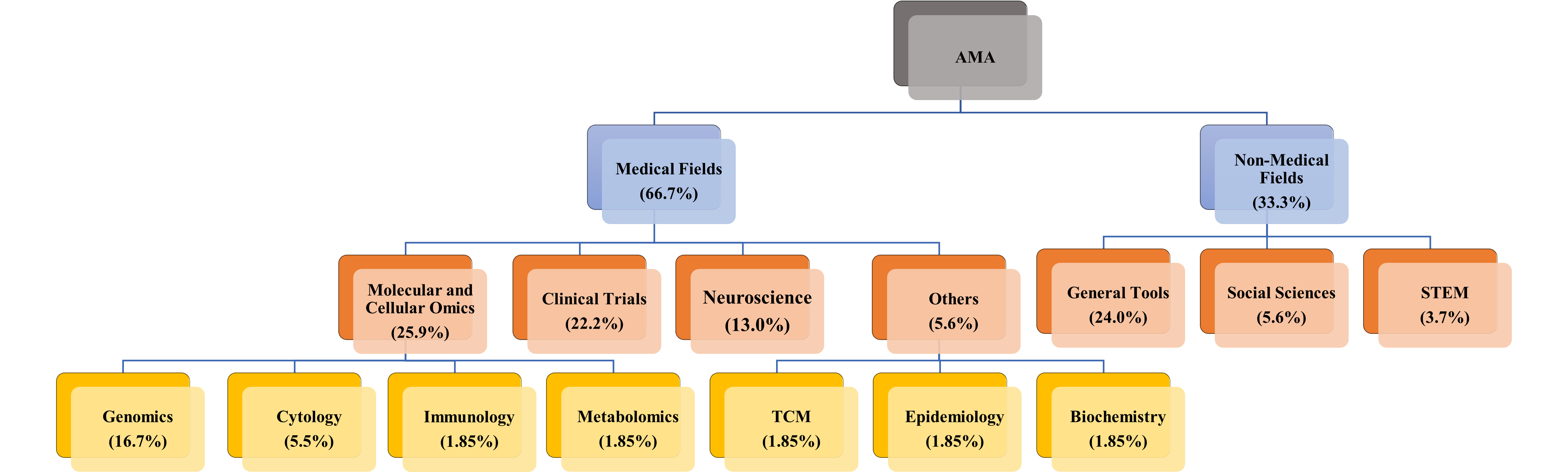}
    \caption{Interdisciplinary applications of AMA across various domains}
    \label{fig:all field}
\end{figure}

\begin{figure}[htbp]
    \centering
    \includegraphics[width=1.0\textwidth]{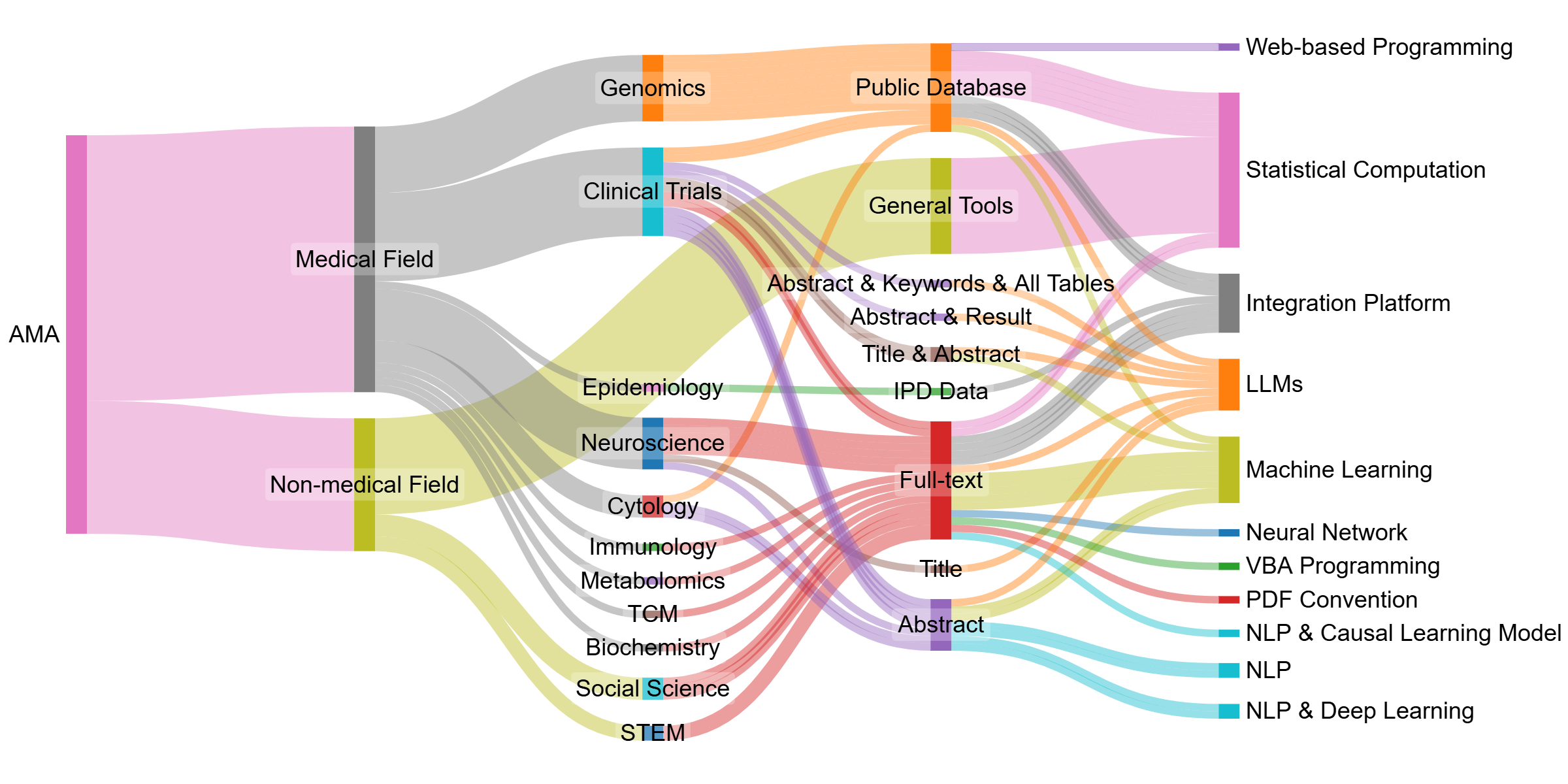}
    \caption{Domain Applications of AMA: Cross-domain Techniques and Data Integration}
    \label{fig:sankey}
\end{figure}

Our systematic review, the first comprehensive examination of AMA across both domains, bridges this knowledge gap by analyzing current technologies, data formats, resolved challenges, and future directions. This dual-perspective approach identifies cross-pollination opportunities, where text-mining techniques from non-medical contexts could enhance medical publication data extraction, while rigorous medical validation methods could inform non-medical research practices. Figures \ref{fig:all field} illustrate AMA's disciplinary breadth, proportional distribution across domains. Specially, Figures \ref{fig:sankey} presents a cross-domain techniques and data integration in AMA, integrating three core dimensions: domain applications (medical vs. non-medical), input data types, and technological methods, with line thickness quantitatively reflecting application frequency. In the medical domain (e.g., clinical trials, neuroscience, immunology), robust linkages emphasize machine learning (ML) and NLP as pivotal technologies for synthesizing structured inputs and unstructured textual evidence (e.g., paper abstracts, full-texts), and LLMs also play a critical role in medical domains. Conversely, the non-medical domain (e.g., STEM, social science) prioritizes ML, deep learning and LLMs from heterogeneous inputs(full-texts) in single lines, which reflect computational scalability in interdisciplinary research, collectively advancing automated, domain-specific AMA.

\subsubsection{Medical field}
AMA in medical filed is rapidly evolving across clinical trials, molecular and cellular omics, neuroscience, and specialized domains. Each field requires domain-specific adaptations of AMA technologies, with distinct patterns of implementation success and challenges from the TTF perspective.

\paragraph{Clinical trials:}
AMA in clinical trials has progressed from abstract-based data extraction to complex full-text analysis, driven by NLP, machine learning, and enhanced computational tools.
\begin{itemize}
    \item \textbf{Specific AMA tasks:} Literature selection, data extraction, publication bias evaluation, results synthesis.
    \item \textbf{Automation tools:} NLP, machine learning, deep learning, and LLMs.
    \item \textbf{Successes:} Foundational result synthesis \cite{michelson_automating_2014}; advanced literature selection \cite{xiong_machine_2018}; improved data extraction efficiency \cite{yang_exploration_2018, pradhan_automatic_2019, mutinda_autometa_2022, mutinda_automatic_2022, shah-mohammadi_large_2024, kartchner_zero-shot_2023, yun_automatically_2024}; publication bias mitigation \cite{lu_cheng_automated_2021}.
    \item \textbf{Challenges:} Data inconsistency and incompleteness; early-stage full-text processing; LLM refinement needs.
    \item \textbf{Domain-specific adaptations:} ChatGPT for radiology abstract screening \cite{issaiy_methodological_2024}; LLM performance in NMA for binary and time-to-event outcomes \cite{reason_artificial_2024}.
\end{itemize}

\paragraph{Molecular and cellular omics:}
Unlike literature-based clinical trials, omics AMA leverages structured datasets from public repositories like Genevestigator \cite{zimmermann_genevestigator_2004}, GEO, and ArrayExpress \cite{parkinson_arrayexpress--public_2004}, shifting focus from data extraction to statistical analysis and integration.
\begin{itemize}
    \item \textbf{Specific AMA tasks:} Data processing, multi-omics integration, and differential expression analysis.
    \item \textbf{Automation tools:} RankProd \cite{hong_rankprod_2006},  \cite{boyko_framework_2016} and \cite{da_devyatkin_towards_2019} for processing gene expression datasets; METAL \cite{willer_metal_2010} for GWAS MA; ShinyMDE \cite{shashirekha_shinymde_2016} for handling differentially expressed gene detection; MetaGWASManager \cite{z_rodriguez-hernandez_metagwasmanager_2024} for large-scale GWAS data; MetaCyto \cite{hu_metacyto_2018} for high-dimensional cytometry data analysis; Amanida \cite{llambrich_amanida_2022} detect study discrepancies. 
    \item \textbf{Successes:} Enhanced data processing and integration; efficient high-dimensional analysis via MetaCyto \cite{hu_metacyto_2018}.
    \item \textbf{Challenges:} Dataset heterogeneity; platform variability; incomplete metabolomics data; ongoing development of integration algorithms.
    \item \textbf{Domain-specific adaptations:} Amanida for metabolomics data gaps \cite{llambrich_amanida_2022}; MetaGWASManager for automated GWAS \cite{z_rodriguez-hernandez_metagwasmanager_2024}.
\end{itemize}

\paragraph{Neuroscience:}
Neuroscience AMA synthesizes brain-related data using NLP, machine learning, and predictive modeling to identify patterns in cognitive and neural states.
\begin{itemize}
    \item \textbf{Specific AMA tasks:} Brain activation mapping, cognitive intervention analysis, and event-related potentials (ERPs) analysis.
    \item \textbf{Automation tools:} Text-mining algorithms, machine learning, NLP, LLMs.
    \item \textbf{Successes:} Evolution of brain mapping tools from NeuroSynth \cite{yarkoni_large-scale_2011} to NeuroQuery \cite{dockes_neuroquery_2020}, Text2Brain \cite{de_bruijne_text2brain_2021}, and Chat2Brain \cite{wei_chat2brain_2023}; enhanced predictive modeling via NPDS 0.9 \cite{craig_bridging_2017}; cognitive intervention repository through CogTale \cite{sabates_cogtale_2021}.
    \item \textbf{Challenges:} Limited data availability; experimental design variability; processing large unstructured text volumes.
    \item \textbf{Domain-specific adaptations:} Probabilistic ERPs literature analysis \cite{donoghue_automated_2022}; neural networks with probabilistic analysis  for mapping text queries to brain activation Chat2Brain \cite{wei_chat2brain_2023}.
\end{itemize}

\paragraph{Specialized domains:}
AMA applications extend to specialized domains including traditional chinese medicine (TCM), epidemiology, and biochemistry, demonstrating adaptability across diverse research contexts.
\begin{itemize}
    \item \textbf{Specific AMA tasks:} Evidence synthesis, data processing, 
    \item \textbf{Automation tools:} Logic regression, machine learning, NLP.
    \item \textbf{Successes:} \cite{lorenz_automatic_2017} Automated logic regression for epidemiological IPD MAs reducing processing time and errors \cite{lorenz_automatic_2017}.
    \item \textbf{Challenges:} Domain-specific systems lack broader applicability.
    \item \textbf{Domain-specific adaptations:} TCM literature synthesis for splenogastric diseases \cite{zhang_construction_2022}; RetroBioCat Database for biocatalysis data exploration \cite{finnigan_retrobiocat_2023}.
\end{itemize}

\subsubsection{Non-medical field}
AMA applications remain limited outside medicine, with only nascent adoption in three key domains: general tools, social sciences, and STEM. This scarcity reflects both challenges and opportunities for expanding AMA beyond medical contexts.

\paragraph{General Tools:}
General tools primarily advance statistical analysis and computational efficiency for AMA, though challenges in tool complexity and data diversity persist.
\begin{itemize}
    \item \textbf{Specific AMA tasks:} Statistical computation, advanced statistical analysis in NMA.
    \item \textbf{Automation tools:} Bayesian random-effects models, graph theory, web-based, decision rules. 
    \item \textbf{Successes:} Statistical packages (\textit{metafor} \cite{viechtbauer_conducting_2010}, Meta-Essentials \cite{suurmond_introduction_2017}, \textit{metamisc} \cite{debray_framework_2019}); NMA tools (\textit{gemtc, pcnetmeta, netmeta}) \cite{neupane_network_2014}; semi-automated systems (ADDIS) \cite{van_valkenhoef_addis_2013}; analytical frameworks for consistency checks, network connectedness, and bias assessment (Bayesian random-effects models \cite{van_valkenhoef_automating_2012, van_valkenhoef_automated_2016, thom_automated_2019}, CINeMA \cite{nikolakopoulou_cinema_2020}, ROB-MEN \cite{chiocchia_semi-automated_2023}); and web platforms (MetaInsight) \cite{owen_metainsight_2019}. Envanced accessibility of complex meta-analytical procedures for researchers without deep statistical expertise.
    \item \textbf{Challenges:} Limited multi-modal data processing; increasing complexity of modern meta-analytical frameworks.
    \item \textbf{Domain-specific adaptations:} Specialized NMA statistical tools such as ADDIS system, CINeMA, ROB-MEN, MetaInsight.
\end{itemize}

\paragraph{Social Science:}
Social sciences have begun adopting AMA tools for synthesizing diverse data types across disciplines like human resource management, psychology and education.
\begin{itemize}
    \item \textbf{Specific AMA tasks:} Data synthesis, predictive modelling.
    \item \textbf{Automation tools:} Bayesian NMA, LLMs.
    \item \textbf{Successes:} MetaBUS enables efficient meta-analysis across large literature volumes \cite{bosco_metabus_2017}; Bayesian NMA opens new possibilities for quantitative analysis \cite{y_liu_gentle_2023}; MetaMate leverages few-shot prompting for data extraction in education \cite{wang_metamate_2024}.
    \item \textbf{Challenges:} Diverse data types complicating automation, particularly for qualitative data and complex models.
    \item \textbf{Domain-specific adaptations:} Bayesian NMA addressing unique statistical challenges in psychology; specialized platforms for social science (MetaBUS) and education (MetaMate)
\end{itemize}

\paragraph{STEM:}
AMA in STEM shows progress in literature retrieval and data extraction but requires more sophisticated tools for comprehensive meta-analysis.
\begin{itemize}
    \item \textbf{Specific AMA tasks:} Literature retrieval, data extraction.
    \item \textbf{Automation tools:} Machine learning-based tools, deep transfer learning.
    \item \textbf{Successes:} Streamlined data extraction tools such as MetaSeer.STEM \cite{neppalli_metaseerstem_2016} from research articles, improving literature analysis efficiency; deep transfer learning systems for literature retrieval \cite{anisienia_research_2021}.
    \item \textbf{Challenges:} Complex data structures; accurate multi-label classification in diverse literature.
    \item \textbf{Domain-specific adaptations:} Deep transfer learning systems for multi-label classification in Information Systems literature.
\end{itemize}

\section{Challenges and Future Potential for AMA}
Despite increasing adoption of AMA techniques, significant challenges remain that must be addressed to realize its full potential for evidence synthesis. To answer RQ4 (What are the critical gaps and future directions for AMA development, and what obstacles need to be addressed to realize its full potential for evidence synthesis?), this section examines key barriers and future directions to enhance AMA’s credibility and utility. These challenges span multiple dimensions: enhancing analytical capabilities while mitigating automation biases; maintaining methodological rigor and transparency; adapting to evolving research technology developments; gaining broader acceptance among stakeholders; and ensuring reliability of synthesized evidence. Table \ref{tab:future} presents a prioritized roadmap for AMA development, categorizing future research directions based on impact and feasibility. This analysis reveals that advancing AMA requires not only technical innovation, but also methodological refinement and strategic implementation approaches to improve its credibility and utility in diverse research contexts.

{
\begin{table}[htbp]
\centering
\caption{Future Research Directions for AMA}
\renewcommand{\arraystretch}{1.5} 
\fontsize{8pt}{10pt}\selectfont
\begin{tabularx}{\textwidth}{p{2cm}X p{2.5cm} p{2.5cm}} 
\toprule
\textbf{Category} & \textbf{Future Trends} & \textbf{Difficulty} & \textbf{Priority} \\
\midrule

\multirow{3}{=}{Advancing Analytical Depth and Balancing Efficiency} 
& Sophisticated analytical components: Development and validation of automated algorithms for sensitivity analysis, heterogeneity assessment (including subgroup analysis), bias evaluation (e.g., publication bias, risk of bias), and NMA specific analyses (e.g., inconsistency detection). 
& Medium \par \textit{Methodological \& Technical} 
& Immediate \par \textit{High Impact \& Validity} \\
\cline{2-4}
& Balancing Automation and Analytical Rigor: Establishing frameworks and best practices to ensure efficient automation does not compromise the depth and methodological rigor of evidence synthesis, requiring human oversight at critical analytical junctures.
& Medium \par \textit{Methodological \& Organizational} 
& Medium \par \textit{Maintain Credibility \& Trust} \\
\cline{2-4}
& Adapting to Diverse Input Types: Creating flexible AMA systems capable of handling diverse data formats (numerical, text, images, raw data), necessitating modular architectures and standardized input interfaces.
& Low \par \textit{Technical} 
& Immediate \par \textit{Broadened Applicability} \\
\midrule

\multirow{3}{=}{Fine-tuning LLMs in AMA} 
& Enhanced Document Analysis: Developing LLMs specifically fine-tuned for analyzing long and complex academic documents, including effective extraction of data from tables, figures, appendices, and supplementary materials, addressing current limitations in context window size and multi-modal data processing.
& Medium \par \textit{Technical \& Data Availability} 
& Immediate \par \textit{Improved Data Completeness} \\
\cline{2-4}
& Transparent LLM Decision-Making: Implementing XAI techniques to enhance the transparency and interpretability of LLM-driven decisions within AMA workflows, fostering expert validation and building trust in automated outputs, particularly in critical domains like healthcare.
& Medium \par \textit{Technical \& Ethical} 
& Immediate \par \textit{Increased Trust \& Adoption} \\
\cline{2-4}
& Robust Benchmarks and Validation: Designing standardized benchmarks and rigorous validation protocols to systematically evaluate the accuracy, reliability, reproducibility, and potential biases of LLM-generated results in AMA, ensuring quality control and facilitating comparative evaluations across different LLM-based tools.
& Medium \par \textit{Methodological \& Community Effort} 
& Medium \par \textit{Quality Assurance \& Comparability} \\
\midrule

{Living AMA} 
& Dynamic and Continuous Updating: Developing fully automated "Living AMA" systems capable of dynamic, ongoing updates as new evidence emerges, requiring robust monitoring pipelines, algorithms for reconciling conflicting data, and effective version control mechanisms, moving beyond static, periodic updates.
& Difficult \par \textit{Technical, Methodological \& Organizational} 
& Long-term \par \textit{Maintain Relevance \& Actionability} \\
\midrule

{Multidisciplinary Collaborations} 
& Fostering Trust and Effective Communication: Building robust multidisciplinary teams encompassing statisticians, computer scientists, domain experts, information specialists, and policymakers, establishing shared goals, standardized workflows, and effective communication channels to overcome disciplinary silos and maximize AMA impact.
& Difficult \par \textit{Organizational \& Social} 
& Long-term \par \textit{Maximize Impact \& Uptake} \\
\midrule

{Interpretability and Transparency in AMA} 
& Establishing XAI Standards and Best Practices: Developing and disseminating standards and best practices for the integration of Explainable AI (XAI) within AMA workflows, focusing on communicating decision-making processes, uncertainty levels, potential limitations, and ensuring responsible and ethical automation.
& Difficult \par \textit{Ethical, Methodological \& Community Effort} 
& Long-term \par \textit{Ethical \& Responsible Innovation} \\
\bottomrule
\end{tabularx}
\label{tab:future}
\end{table}
}

\subsection{Advancing Analytical Depth and Balancing Efficiency in AMA}
A critical and persistent limitation in AMA remains the automation of advanced analytical methodologies, including sensitivity analyses, heterogeneity assessments, publication bias evaluations, and stratified subgroup analyses. While preliminary data processing has advanced significantly, sophisticated analytical automation remains underexplored, compromising reproducibility and scientific validity of AMA findings. Future research should prioritize three critical areas: (1) Algorithm advancement. Developing frameworks that execute complex analytical functions with minimal human intervention while maintaining methodological rigor, including automated sensitivity analysis and bias detection tools. (2) Methodological balance. Creating frameworks that enhance efficiency without compromising analytical depth and integrity, with strategic human oversight at critical analytical stages. (3) Multi-modal data integration. Incorporating heterogeneous data types (numerical data, medical images, tables, raw data) through adaptable extraction techniques for comprehensive, statistically sound evidence synthesis. These advancements would elevate AMA beyond basic automation to deliver both sophisticated analytical capabilities and enhanced efficiency, strengthening its credibility in high-impact research domains.

\subsection{Fine-turning LLMs in AMA}
LLMs offer transformative potential for AMA by efficiently processing unstructured text and extracting critical variables (effect sizes, confidence intervals) from research articles. However, several challenges hinder their full-scale deployment: hallucinations that fabricate results, unacceptable in high-stakes applications like healthcare; propagation of implicit biases from training corpora into synthesized outputs; and limitations with extensive context windows when processing journal articles, dissertations, and complex figures/tables \cite{liu_lost_2024}. To maximize LLM utility in AMA, future research should prioritize: (1) Developing models capable of effectively analyzing long, complex documents and extracting data from tables and figures in academic articles. (2) Enhancing transparency through explainable AI (XAI) techniques to facilitate expert validation of automated extractions. (3) Designing benchmarks and protocols to ensure the accuracy, reliability, and reproducibility of LLM-generated results. These advancements will significantly enhance the reliability and interpretability of LLM-assisted AMA workflows for evidence synthesis. 

\subsection{Living AMA}
Current AMA implementations primarily automate discrete stages of MA but lack mechanisms for continuous, real-time evidence updates. This limitation is particularly evident in Cochrane MAs, which require periodic updates to maintain clinical relevance. A "living AMA" addresses this gap by envisioning a system that can automatically and continuously scan databases for new studies, extract relevant data, and integrate fresh evidence into existing analyses. Realizing this vision should focus on three key aspects. First, designing robust AI-driven mechanisms to identify and validate new studies as they emerge. Second, developing algorithms to make a version control and reconcile conflicting data across studies while preserving analytical transparency. Third, creating efficient alert mechanisms that update researchers without overwhelming them with excessive information. Living AMA approaches have already emerged in related domains, such as "living literature review" \cite{wijkstra_living_2021}, COVID-19 living MAs \cite{cochrane_emergency_and_critical_care_group_interventions_2020}, MetaCOVID project \cite{evrenoglou_metacovid_2023} and SOLES system \cite{hair_systematic_2023}. Building on these foundations, future work must refine the methodological framework for Living AMA to ensure delivery of up-to-date, high-quality evidence synthesis.

\subsection{Fostering Multidisciplinary Collaborations}
The success of AMA depends on requiring seamless collaborations between statisticians, computer scientists, domain experts, and policymakers. However, interdisciplinary cooperation remains a bottleneck due to differences in methodologies, terminology, and research priorities. Addressing this challenge requires three strategic approaches: (1) interdisciplinary training programs to familiarize researchers with AMA methodologies and computational techniques; (2) joint funding initiatives to support large-scale, collaborative AMA projects; (3) shared platforms and community to promote cross-disciplinary integration. These approaches leverage complementary expertise: statisticians ensure methodological rigor, computer scientists develop the technical framework, and domain experts provide contextual knowledge to interpret findings meaningfully. Through effective communication and trust-building, AMA can evolve into a widely adopted tool bridging computational power with domain-specific expertise.

\subsection{Interpretability and Transparency in AMA}
As AMA tools become more sophisticated, transparency in their decision-making processes becomes increasingly paramount, particularly in high-stakes domains such as medical research where evidence synthesis directly influences clinical decisions. The integration of explainable AI (XAI) methods into AMA represents a critical frontier in ensuring credibility and adoption. The challenge of interpretability in AMA extends beyond mere technical performance. While automated systems can significantly reduce the time and effort required for MA, their value diminishes if end-users cannot understand or trust their outputs. This is particularly crucial during the evidence synthesis phase, where complex algorithms process and integrate diverse evidence sources. Recent research \cite{luo_evaluating_2024} highlights the delicate balance required between efficient automation and maintaining the depth and accuracy of evidence synthesis. Future research should prioritize the standardization of XAI integration within AMA workflows, ensuring automated processes remain transparent, reproducible, and trustworthy. Various XAI techniques such as rule-based explanations, visual explanations, sensitivity analysis may integrate into AMA findings with more accessible and easier adjustments. Through these approaches, AMA can evolve into robust and widely accepted tools that enhance the quality of evidence synthesis.

\section{Discussion}
AMA has emerged as a transformative innovation in quantitative evidence synthesis, driven by exponential growth in literature that demands efficient, scalable, and reproducible quantitative research methods. Advanced AI, particularly 'thinking models' with the capable of complex reasoning, has become a cornerstone of this evolution. This review has provided an evaluation of AMA via a descriptive lens (RQ1), analytical lens (RQ2), comparative lens (RQ3) and a future-oriented lens (RQ4). Despite AMA offering significant benefits compared to traditional MA, full automation remains aspirational rather than becoming a standard. This gap underscores the urgent imperative to harness 'thinking models', bridging technical and methodological barriers to position AMA as a critical frontier fro future evidence synthesis innovation. 

\subsection{Methodological Disparities Between CMA and NMA}
Quantitative analysis reveals a clear research imbalance, with 81\% of AMA studies focusing on CMA versus only 19\% addressing NMA. This disparity arises from the inherent complexity of NMA, which requires integrating both direct and indirect comparisons across multiple interventions while accounting for data heterogeneity. CMA, involving primarily pairwise comparisons, is more amenable to automation through established statistical frameworks. Table \ref{tab:featuresnma} highlights key distinctions between CMA and NMA automation. While CMA utilizes standard statistical models (fixed-effects, random-effects) automated through established software packages such as \textit{metaMA} \cite{marot_moderated_2009} and \textit{metafor} \cite{viechtbauer_conducting_2010}, NMA requires more sophisticated Bayesian or frequentist models. Visualization in CMA focuses on forest and funnel plots, whereas NMA demands complex network graphs and inconsistency plots, often requiring manual adjustments. Addressing the NMA automation gap necessitates algorithms capable of handling multi-dimensional data while ensuring model transparency.

{
\begin{table}[hbt]
    \caption{Comparative Analysis of CMA and NMA Automation}
    \centering
    \fontsize{8pt}{10pt}\selectfont
    \begin{tabular}{p{3cm}p{5cm}p{5cm}}
        \toprule
        \textbf{Feature} & \textbf{Conventional Meta-analysis (CMA)} & \textbf{Network Meta-analysis (NMA)} \\
        \midrule
        Data Structure & Homogeneous data, single pairwise comparisons & Heterogeneous data, multiple interventions, network structure \\
        \midrule
        Statistical Models & Fixed-effects, random-effects models & Bayesian models, frequentist network approaches \\
        \midrule
        Automation Tools & \texttt{metaMA}, \texttt{metafor}, NLP, LLMs & \texttt{gemtc}, \texttt{netmeta}, network tools \\
        \midrule
        Visualization & Forest plots, funnel plots & Network graphs, inconsistency plots \\
        \midrule
        Automation Success & \textbf{High} in data extraction, but \textbf{low} in synthesis & \textbf{Moderate}; challenges in network construction and inconsistency analysis \\
        \bottomrule
    \end{tabular}
    \label{tab:featuresnma}
\end{table}
}

\subsection{Complexity of Full Automation}
Despite notable advancements, full AMA workflow remains elusive. Our review reveals that majority of existing studies employed semi-automated approaches, with automation largely confined to data extraction and preliminary synthesis. Only one study \cite{yang_exploration_2018} has explored full automation across all AMA stages. This highlights a critical gap between the automation of individual components, highlighting the gap between automating individual components and developing integrated, end-to-end systems. Key barriers include: (1) Technical challenges. Data heterogeneity across formats (structured databases, unstructured literature, high-dimensional biomedical data); computational complexity of advanced meta-analytic methods; and LLM limitations in interpreting context-sensitive statistical details. (2) Methodological barriers.  Difficulty automating qualitative judgments (risk-of-bias assessment, confounding adjustment, evidence grading). (3) Organizational and infrastructural hinders. Limited cross-disciplinary adaptability and absence of universal standards for seamless data integration. Addressing these challenges requires a multi-faceted approach that combines advancements in AI, robust methodological frameworks, and domain expertise to ensure both automation feasibility and scientific validity.

\subsection{Ethical and Practical Considerations}
AMA adoption also raises critical ethical questions. One of the most pressing ethical concerns is the risk of bias amplification. AMA systems typically rely on existing datasets, including published literature, trained on existing literature may reinforce systematic publication biases, particularly if they rely on biased data sources. Additionally, the increasing reliance on automation in evidence synthesis introduces concerns about deskilling of researchers, which means as automation takes over certain tasks, researchers may become less proficient in critical appraisal and statistical analysis, potentially reducing the quality of evidence synthesis. Furthermore, the development and adoption of AMA tools are disproportionate, creating risks of global inequity in access to advanced evidence synthesis technologies. If AMA tools remain proprietary, cost-prohibitive, or require specialized technical expertise, low-resource settings may struggle to leverage these innovations, potentially widening disparities in research capacity. Another ethical use of AMA is transparency, without transparency, stakeholder trust in automated evidence synthesis may be undermined, raising concerns about reproducibility and accountability in decision-making. Therefore, researchers looking to adopt AMA should consider that: (1) not all AMA tools are equally effective across disciplines. Choosing the right tool requires an understanding of its strengths, limitations, and adaptability. (2) Rather than seeking full automation, researchers should integrate AMA as an assistive tool while maintaining expert oversight in critical analytical processes. Through these approaches, researchers can balance ethical responsibility, methodological rigor, and transparency without overshadow AMA potential benefits.  

\subsection{Implications for Evidence Synthesis}
The ability of AMA to streamline quantitative evidence synthesis has been widely recognized across biomedicine, neuroscience, epidemiology, and omics research through automating data extraction, statistical modeling, and synthesis processes. Its evolution could significantly enhance efficiency, reproducibility, and scalability. For example, the continued advancement of AMA has the potential to reshape the landscape of evidence synthesis, which enabling more dynamic and responsive updates to existing evidence bases. This could be particularly valuable in rapidly evolving research domains, such as pandemic response or emerging medical technologies. Besides, automation approaches can facilitate data extraction and statistical analysis, thereby minimizing inconsistencies introduced by subjective human interpretation. Large-scale and complex analysis AMA from heterogeneous datasets will extend beyond traditional systematic reviews based solely on published clinical trials.

However, over-reliance on automation without addressing limitations risks undermining synthesized evidence reliability. One of the concerns is diminished role of expert judgment in study selection, data interpretation, and result contextualization. Besides, many AMA tools operate as black-box systems, making it challenging to trace how decisions—such as study inclusion/exclusion criteria—are made. Moreover, if automation is trained on biased or incomplete datasets, it may affect the accuracy of evidence synthesis, thereby affecting clinical and policy-related decision-making. Furthermore, most existing AMA systems primarily rely on published studies indexed in databases such as PubMed or Scopus, this focus may reinforce existing publication biases by systematically underrepresenting negative or inconclusive findings, particularly those available only in gray literature, preprints, or non-English sources. Addressing this issue will require developing more inclusive and adaptive search strategies within AMA frameworks.

The emergence of 'thinking models' with complex reasoning in advanced AI and LLMs presents a transformative opportunity to revolutionize AMA by bridging computational power with sensitive analytical capabilities. They enable adaptive analytical strategies that can dynamically handle multi-modal datasets, reducing human intervention while maintaining methodological precision. To maximize the benefits of AMA, a balanced and methodologically rigorous approach is therefore essential, integrating 'thinking LLMs' while mitigating its inherent challenges. Domain experts should remain actively involved in tasks such as study selection, risk of bias assessment, sensitivity analyses, and interpretation, ensuring AMA outputs align with established research methodologies. Standardized reporting frameworks should be established to enhance the transparency of AMA methodologies, allowing researchers to audit and validate automated results. More sophisticated statistical modeling techniques and advanced AI techniques, should be developed based on data complexity. Finally, as AMA tools become more widely adopted, policymakers should establish clear guidelines in evidence synthesis such as ethical considerations in AI-driven MAs and equitable access to AMA technologies. By integrating these principles and embracing AI-driven breakthroughs, AMA can evolve into a more robust and ethically responsible tool for evidence synthesis, bridging the gap between automation-driven efficiency and the need for methodological rigor and interpretability, and fundamentally transform evidence synthesis across disciplines. 

\subsection{Study Limitations}
This review is constrained by several interconnected challenges. First, this study's predominant focus on well-documented tools from literature databases, potentially overlooks other innovative methodologies. Moreover, given the rapid evolution of automation technologies, particularly in artificial intelligence and LLMs, the review's findings may quickly become outdated without regular updates. The dynamic nature of this field necessitates continuous revision to maintain relevance and usefulness. Second, while the proposed PPS with the TTF model offers a framework to understand AMA development, the criteria for assessing the level of automation remain subjective and qualitative, making it difficult to quantitatively compare the automation capabilities of different tools. Developing more standardized criteria for evaluating these tools would enhance the objectivity and reliability of future reviews. Third, this review is primarily based on the summary and classification of existing literature, without conducting empirical validation or performance evaluation (e.g. comparative experiments of different tools). As a result, some conclusions rely on self-reported findings in the reviewed studies, lacking independent external verification. These limitations highlight the need for ongoing research and development to refine AMA tools, address integration challenges, and ensure that they remain reliable and applicable in diverse research contexts.

\section{Conclusion}
Meta-analyses are critical to advancing science.
The prospect of automating meta-analyses opens opportunities for transforming quantitative research synthesis and redefining scientific progress. The automation of meta-analyses is desperately needed right now to manage the expanding volume of academic research.
We currently stand at the threshold of a significant AI revolution which holds potential to provide solutions to many remaining limitations and unsolved questions in the field of meta-analysis automation. To proceed effectively and maximize this opportunity, this study fills the gap in literature by comprehensively investigating the current landscape of these automation efforts for meta-analyses using a robust framework. This research has assessed existing methodological approaches, compared implementation patterns across various domains, and synthesized key challenges as well as future directions. Our emphasis has been on the potential that increasingly sophisticated large language models with enhanced reasoning capabilities offer to accelerate progress further. 
Our research has found that automated tools have excelled at streamlining data extraction and statistical modeling, yet they still remain limited in achieving full-process automation, particularly in advanced synthesis and bias evaluation. Our work finds that future research efforts must prioritize the development of integrated frameworks that not only enhance individual meta-analytic stages but also bridge gaps between them. Efforts need to focus on also refining AI-driven models to improve interpretability and robustness, ensuring that heterogeneous data sources and complex synthesis tasks are effectively managed. Furthermore, standardizing methodologies across disciplines will be essential to unlock the full transformative potential of automated meta-analysis.

Therefore, as the volume and complexity of academic research continue to escalate, the evolution of automated meta-analysis represents a pivotal innovation for evidence synthesis. By harnessing advanced AI capabilities and addressing current methodological shortcomings, the research community can significantly enhance the efficiency, accuracy, and reproducibility of meta-analytic practices—ultimately revolutionizing the way we synthesize scientific knowledge.

\bibliographystyle{unsrtnat}


\clearpage

    
\appendix
\renewcommand{\thetable}{Appendix~\arabic{table}} 
\setcounter{table}{0}
\captionsetup[table]{labelformat=empty}
\section*{Appendix}

\ref{tab:all studies} Key Insights from Studies on Automated Meta-Analysis.
\fontsize{7pt}{10pt}\selectfont
\begin{longtable}{p{0.3cm}p{0.3cm}p{2.2cm}p{0.3cm}p{1.0cm}p{5.5cm}p{1.0cm}p{1.0cm}p{1.0cm}}
    \caption{Key Insights from Studies on Automated Meta-Analysis} 
    \label{tab:all studies} \\ \hline
    \multirow{2}{*}{\textbf{No}} & 
    \multirow{2}{*}{\textbf{Ref}} & 
    \multirow{2}{*}{\textbf{Core Contribution}} & 
    \multirow{2}{*}{\textbf{Year}} & 
    \multirow{2}{*}{\textbf{Field}} & 
    \multirow{2}{*}{\textbf{Methodology \& Purposes}} & 
    \multicolumn{3}{c}{\textbf{Automated Processing Steps}} \\ \cline{7-9}
    & & & & & & \textbf{Pre-processing} & \textbf{Processing} & \textbf{Post-processing} \\ \hline
    \endfirsthead

    \hline
    \multirow{2}{*}{\textbf{No}} & 
    \multirow{2}{*}{\textbf{Ref}} & 
    \multirow{2}{*}{\textbf{Core Contribution}} & 
    \multirow{2}{*}{\textbf{Year}} & 
    \multirow{2}{*}{\textbf{Field}} & 
    \multirow{2}{*}{\textbf{Methodology  \& Purposes}} & 
    \multicolumn{3}{c}{\textbf{Processing Steps}} \\ \cline{7-9}
    & & & & & & \textbf{Pre-} & \textbf{Processing} & \textbf{Post-} \\ 
    & & & & & & \textbf{processing} & & \textbf{processing} \\ \hline
    \endhead

    \hline
    \multicolumn{9}{r}{\textit{Table continues on the next page...}} \\ \hline
    \endfoot

    \hline
    \endlastfoot

    1&\cite{hong_rankprod_2006}& RankProd	&2006&	Molecular and Cellular Omics &	Mathematical calculations&	&※	&※ \\ \hline
    2&\cite{choi_latent_2007}&metaArray	&2007&	Molecular and Cellular Omics &	two general probabilistic models for quantities that are combinable across studies:(1) Markov Chain Monte Carlo techniques (2) expectation-maximization algorithm&		&※	&
    \\ \hline
    3& \cite{marot_moderated_2009}&metaMA	&2009 &Molecular and Cellular Omics &Mathematical in effect size calculation and P-value combination	&	&※	&
    \\ \hline
    4& \cite{viechtbauer_conducting_2010}&metafor	&2010	&General Tools&	Mathematical calculations&	&	※	&
    \\ \hline
    5& \cite{willer_metal_2010}&METAL&	2010 &Molecular and Cellular Omics &computationally efficient tool for MA of genome-wide association scans: (1) converts the direction of effect and P-value observed in each study into a signed Z-score (2) weights the effect size estimates, or $\beta$ -coefficients, by their estimated standard errors	&	&※	&
    \\ \hline
    6	&\cite{yarkoni_large-scale_2011}&
    NeuroSynth	&2011&	Neuroscience&	Identified psychological terms, extracted brain activation coordinates, mapped brain-cognition links, and decoded constructs using naïve Bayes classification.	&	&	&※
    \\ \hline
    7	&\cite{feichtinger_cancerma_2012}&
    CancerMA	&2012	&Molecular and Cellular Omics&	Raw data from ArrayExpress and GEO repositories underwent manual assessment, quality control with the 'simpleaffy' R package, and pre-processing using standard methods. The data was annotated with databases like Ensembl and HGNC, integrated into the CancerMA website, analyzed using the 'GOstats' R package, and visualized.	&	&	&※
    \\ \hline
    8&	\cite{wang_r_2012}&
    MetaOmics	&2012&	Molecular and Cellular Omics&	Mathematical calculations	&	&※	&
    \\ \hline
    9&	\cite{van_valkenhoef_automating_2012}&
    Bayesian model&	2012&General Tools&	Bayesian homogeneous variance random effects consistency models	&	&※	&
    \\ \hline
    10&	\cite{van_valkenhoef_addis_2013}&
    ADDIS	&2013&	Clinical Trials	&The unifying data model together with a semi-automated analysis generation system&		&	&※
    \\ \hline
    11	&\cite{j_feichtinger_cancerest_2014}&
    CancerEST&	2014	&Molecular and Cellular Omics	&obtained the complete data available from the Unigene database, set up a local MySQL database and examined expression profile of submitted genes	&	&	&※
    \\ \hline
    12	&\cite{michelson_automating_2014}&
    First attempt of AMA in RCTs	&2014&	Clinical Trials&	Transformed clinical trial abstracts into structured data, groups studies with similar treatments and outcomes, and uses the Paule-Mandel random-effects model to determine the overall treatment effect	&※&	※&	※
    \\ \hline
    13	&\cite{neupane_network_2014}&
    gemtc, pcnetmeta, netmeta&	2014&	General Tools&	Mathematical calculations&	※&	※	&※
    \\ \hline
    14	&\cite{boyko_framework_2016}& AMA specially in Dendritic Cell Therapy	&2016 &Molecular and Cellular Omics&	Patient-level data on dendritic cell vaccination from 71 studies was semi-supervisedly extracted and transformed into a vector space for analysis and classification.&	&	※	&
    \\ \hline
    15	&\cite{neppalli_metaseerstem_2016}&
    MetaSeer.STEM&	2016	&STEM	&automatically converted the PDF into text and manually mark the data of interest, then dataset is used to train classifiers in supervised learning &		&※	&
    \\ \hline
    16	&\cite{shashirekha_shinymde_2016}&
    ShinyMDE&	2016&	Molecular and Cellular Omics &	multiple statistical methods, including Fisher’s, Stouffer’s, Minimum p-value, and Maximum p-value, along with their one-sided correction variants	&	&	※&
    \\ \hline
    17&	\cite{van_valkenhoef_automated_2016}&
    Node-splitting model&	2016	&General Tools	&consistency models used in priors and starting values for node-splitting models &		&※	&
    \\ \hline
    18	&\cite{bosco_metabus_2017}&
    MetaBUS	&2017&	Social Science	&A matrix is extracted via OCR, cleaned with VBA scripts, reviewed for errors, and transposed into a standardized format.&	※	&	&
    \\ \hline
    19&	\cite{craig_bridging_2017}&
    Nexus-PORTAL-DOORS System v0.9	&2017&	Neuroscience	&system with components for metadata curation, data retrieval, natural language processing, query expansion, inference extraction, and statistical analysis	&	&	&※
    \\ \hline
    20	&\cite{lorenz_automatic_2017}&
    PROG-IMT&	2017	&Epidemiology&	Manual rules were created for each target variable, optimized with ROC analysis, and used in Boolean logic regression. The simulated annealing algorithm identified the best rule combination to minimize errors.	&	&※	&
    \\ \hline
    21&	\cite{suurmond_introduction_2017}&
    Meta‐Essentials	&2017&	General	Tools&a set of 7 workbooks each designed to serve a special purpose. Each workbook consists of 6 sheets.
    Mathematical calculations	&	&	&※
    \\ \hline
    22	&\cite{hu_metacyto_2018}&
    MetaCyto	&2018&	Molecular and Cellular Omics&unsupervised analysis and guided analysis to identify common cell subsets	&	&	&※
    \\ \hline
    23	&\cite{xiong_machine_2018}&
    robust automatic study selection	&2018&	Clinical Trials	&Publications were grouped using K-means, and relevant clusters identified with maximum entropy classification, followed by manual screening	&※		& &
    \\ \hline
    24	&\cite{yang_exploration_2018}&
    Automating processes in MA through computer technology&	2018&	Clinical Trials	&Screened in the URL format of PubMed + PMID +XML, extracted table information through smallPDF and analyzed results in R&	※&	※	&※
    \\ \hline
    25	&\cite{da_devyatkin_towards_2019}&
    text analysis in cell-based immunotherapy&	2019&	Molecular and Cellular Omics	&The framework crawled PubMed abstracts, extracted features, identified entities, analyzed relationships, filtered uninformative entities, and mined co-occurrence statistics	&	&※	&
    \\ \hline
    26	&\cite{debray_framework_2019}&
    metamisc&	2019&	Clinical Trials	&Mathematical calculations	&	&※	&
    \\ \hline
    27&	\cite{owen_metainsight_2019}&
    MetaInsight	&2019&	General Tools	&used R's netmeta and Shiny packages for analysis and user interface	&	&	&※
    \\ \hline
    28&	\cite{pradhan_automatic_2019}&
    EXACT	&2019&	Clinical Trials	&Python library for parsing ClinicalTrials.gov records and an interface for users to specify desired data	&	&※	&
    \\ \hline
    29	&\cite{thom_automated_2019}&
    Stata package”network”&	2019&	General Tools	&Used graph theory to assess the connectivity of evidence networks in network meta-analysis by constructing an adjacency matrix, where rows and columns represent treatments, and entries indicate whether treatments were compared, with zeros on the diagonal.	&	&※&	
    \\ \hline
    30	&\cite{dockes_neuroquery_2020}&
    NeuroQuery	&2020	&Neuroscience	&Used a multivariate model trained on 13,459 publications, and inferred semantic similarities across terms using NLP	&	&※&	
    \\ \hline
    31	&\cite{nikolakopoulou_cinema_2020}&
    CINeMA&	2020&	General Tools	&Mathematical calculations&		&	&※
    \\ \hline
    32	&\cite{penaloza_towards_2020}&
    first approach for automated reasoning in meta-analyses	&2020&	General Tools&	Mathematical calculations&		&	&※
    \\ \hline
    33	&\cite{alisa_method_2021}&
    Information extraction of immunosuppressive cell&	2021	&Molecular and Cellular Omics	&hybrid approach using dictionaries, a rule-based parser, and a pre-trained machine learning model to identify and filter entities, leveraging external linguistic resources and syntactic-semantic features for entity relationships	&	&※	&
    \\ \hline
    34	&\cite{anisienia_research_2021}&
    Help authors to automate parts of a literature review and mitigate some of the problems associated with everincreasing number of papers	&2021&	STEM	&Used deep transfer learning methods for multi-label classification in recognizing research methods&	※	&	&
    \\ \hline
    35	&\cite{lu_cheng_automated_2021}&
    AMA in causal learning perspective&	2021&	Clinical Trials	&two-step framework: (1) automatic data extraction from publications using NLP, and (2) automatic inference of treatment effects, controlling for biases.&		&※	&※
    \\ \hline
    36	&\cite{de_bruijne_text2brain_2021}&
    Text2Brain	&2021&	Neuroscience&	consists of a transformer-based text encoder and a 3D CNN&		&	&※
    \\ \hline
    37&	\cite{sabates_cogtale_2021}&
    CogTale	&2021	&Neuroscience&	three-tier web application with a React-based user interface, a NodeJS backend with a REST API, and an R-based analysis sub-module for statistical analysis and report generation. It uses MongoDB for data storage and integrates with a WordPress site for public access	&	&	&※
    \\ \hline
    38	&\cite{donoghue_automated_2022}&
    automated meta-analytic tool for ERP-related literature	&2022	&Neuroscience	&used search terms to gather articles and create data-driven profiles of ERP components, all literature data was collected using the E-utilities API	&	&※	&
    \\ \hline
    39	&\cite{llambrich_amanida_2022}&
    Amanida	&2022	&Molecular and Cellular Omics&	Mathematical calculations	&	&※&	
    \\ \hline
    40	&\cite{mutinda_automatic_2022}&
    PICO recognition&	2022	&Clinical Trials&	developed a BERT-based NER model to extract PICO information from abstracts of breast cancer	&	&※&	
    \\ \hline
    41	&\cite{mutinda_autometa_2022}&
    AUTOMETA	&2022	&Clinical Trials	&developed a BERT-based to extract PICO information from abstracts&		&※	&
    \\ \hline
    42	&\cite{zhang_construction_2022}&
    developed and tested a new form of clinical evidence in TCM	&2022&	TCM	&created a data sheet for extraction based on Microsoft Excel powered with Visual Basic for Applications, developed TCM database with Python 3.8 and MySQL 8	&	&※&	
    \\ \hline
    43	&\cite{chiocchia_semi-automated_2023}&
    ROB-MEN web-application	&2023	&General Tools	&Integrated with CINeMA framework, semi-automated some of the required steps of the ROB-MEN tool and produced the two output tables in a ready-to-use .csv format	&	&	&※
    \\ \hline
    44	&\cite{finnigan_retrobiocat_2023}&
    RetroBioCat&	2023	&Biochemistry	&a python Flask web server that employs Jinja2 to render HTML pages, utilizing Bootstrap 4 and custom Javascript to provide the user interface, the Scikit-learn, Pandas, Biopython, and NumPy python packages were employed 	&	&	&※
    \\ \hline
    45	&\cite{kartchner_zero-shot_2023}&
    Zero-shot Extraction	&2023&	Clinical Trials	&Two models, GPT-3.5 Turbo and GPT-JT, were selected. Prompts were created through an iterative process with database curators, often including value lists for multiple-choice questions	&	&※	&
    \\ \hline
    46	&\cite{wei_chat2brain_2023}&
    Chat2Brain	&2023&	Neuroscience	&used LLMs (ChatGPT) and a text-to-image model to map text queries to brain activation map	&※	&	&※
    \\ \hline
    47&	\cite{y_liu_gentle_2023}&
    BUGSnet	&2023	&Social Science	&automated R package (Bayesian inference Using Gibbs Sampling)	&	&	&※
    \\ \hline
    48	&\cite{issaiy_methodological_2024}&
    Evaluate ChatGPT's performance&	2024	&Clinical Trials	&compare ChatGPT's performance in screening radiology abstracts with general physicians using sensitivity, specificity, PPV, NPV, and workload saving&	※	&	&
    \\ \hline
    49&	\cite{luo_evaluating_2024}&
    Evaluate the Efficacy of LLMs in MA	&2024&	General Tools&	evaluated efficacy of LLMs by comparing with expert reviews using statistical methods, analyzing metrics like accuracy, sensitivity, specificity, predictive values, F1-score, and Matthews correlation coefficient	&※	&	&
    \\ \hline
    50	&\cite{reason_artificial_2024}&
    Evaluate LLMs in NMA	&2024	&Clinical Trials	&Four case studies were used to develop a Python script that utilizes an LLMs to automate data extraction, NMA script generation, and report creation	&	&	&※
    \\ \hline
    51	&\cite{shah-mohammadi_large_2024}&
    LLMs in data extraction of AMA&	2024	&Clinical Trials	&used the NCBI API to retrieve clinical trial papers, extracting key data from the XML content, and employing GPT to generate SQL queries for structuring and analyzing the data and saved as a CSV for further analysis	&	&※	&
    \\ \hline
    52&	\cite{wang_metamate_2024}&
    MetaMate	&2024&	Social Science&	Few-shot prompting for in-context learning	&	&※	&
    \\ \hline
    53&	\cite{yun_automatically_2024}&
    Annotated a modest but granular evaluation dataset of RCTs	&2024&	Clinical Trials&	Zero shot in LLMs	&	&※	&
    \\ \hline
    54&	\cite{z_rodriguez-hernandez_metagwasmanager_2024}&
    metaGWASmanager	&2024&	Molecular and Cellular Omics	&based on R, Bash, and Python, involved customizing analysis scripts, prepared and validated phenotype and genotypic data, conducted GWAS, performed quality control, and finalized the meta-analysis using METAL.	&	&	&※
\end{longtable}


\end{document}